%% file: main.tex
\documentclass[runningheads]{llncs}

\usepackage{eccv}

\usepackage{eccvabbrv}

\usepackage{graphicx}
\usepackage{subcaption}
\usepackage{booktabs}

\usepackage{makecell}
\usepackage{multirow}
\usepackage{hhline} 
\usepackage{siunitx}
\usepackage{etoolbox}
\newcommand{\boldres}[1]{{\fontseries{b}\selectfont #1}}
\robustify\boldres

\usepackage{array}
\usepackage{microtype}   
\usepackage{ragged2e}

\usepackage{algorithm}
\usepackage{algpseudocode} %
\algrenewcommand\algorithmicrequire{\boldres{Input:}}
\algrenewcommand\algorithmicensure{\boldres{Output:}}
\algtext*{EndIf}
\algtext*{EndWhile}
\algtext*{EndFor}

\usepackage[accsupp]{axessibility}  

\usepackage[pagebackref]{hyperref}

\usepackage{orcidlink}

\begin{document}

\title{REVA-PO: Stabilizing Reinforcement Learning for Chest X-ray Report Generation} 

\titlerunning{REVA-PO for CXR Report Generation}

\author{Li Guo \and
Anas M. Tahir \and
Z. Jane Wang}

\authorrunning{L.~Guo et al.}

\institute{University of British Columbia, Vancouver, BC, Canada \\
\email{lguo@ece.ubc.ca}}

\maketitle
\begin{center}
    \textit{Accepted to ECCV 2026}
\end{center}

\begin{abstract}
  Automated chest X-ray report generation has recently benefited from reinforcement learning (RL) and large language models. However, RL training often suffers from instability or limited exploration due to fixed Kullback-Leibler (KL) regularization and a static reference policy that accumulates KL pressure over time. We propose Response-Weighted and Validation-Anchored Policy Optimization (REVA-PO), a RL framework that stabilizes long-term training via Response-Weighted Regularization (RER) and Validation-Anchored Policy Reset (VAPR). RER dynamically adjusts per-response KL weights based on advantage and reference-policy entropy, relaxing constraints for high-quality responses while tightening them for low-quality ones. Complementarily, VAPR periodically synchronizes the reference and current policies to the best validation checkpoint, resetting accumulated regularization pressure to expand the viable exploration space. To ensure a robust starting point, we employ a three-stage pipeline consisting of warm-up training, classifier-guided supervised fine-tuning, and RL. Extensive evaluations on MIMIC-CXR and IU-Xray demonstrate that REVA-PO sets new state-of-the-art benchmarks in both linguistic quality and clinical accuracy. Notably, BLEU-4 improves by 5.1\% on MIMIC-CXR and 3.6\% on IU-Xray, while CheXpert F1 and RadGraph F1 scores increase by 4.5\% and 12.8\%, respectively, over prior leading methods. The code is publicly available at \url{https://github.com/LiGuo12/REVA_PO/}.
\end{abstract}

\section{Introduction}
\label{sec:intro}

Chest X-rays (CXR) are essential for diagnosing and monitoring chest organ conditions, yet the demand for interpretation often exceeds radiologist capacity, increasing turnaround times and diagnostic risks \cite{rimmer2017radiologist, krupinski2010long, berlin2000liability}. While automated CXR report generation (CXR-RG) has progressed significantly \cite{sloan2024automated, guo2024automatic}, generated reports still struggle with nuanced word choice and sentence fluency compared to professional reports.

To improve report quality, recent work has integrated large language models (LLMs) \cite{liu2024bootstrapping, jin2025chain, yang2025spatio,wang2025cxpmrg, wang2024hergen}. Standard approaches often utilize contrastive learning to map radiographic features into LLM embedding space—a process that is memory-intensive and prone to overfitting on limited medical datasets. While the emerging large vision-language models (LVLMs) pretrained on broad image–text data already align visual and textual spaces \cite{bai2025qwen2, grattafiori2024llama, xu2025llava}. Inspired by the advantages of LVLMs, we propose adopting a pretrained LVLM for CXR-RG, which eliminates the need to train a separate image encoder and reduces memory use while mitigating overfitting.

From a training perspective, reinforcement learning from human feedback (RLHF) has demonstrated superior performance in text generation \cite{ouyang2022training, bai2022training, ziegler2019fine}. Standard proximal policy optimization (PPO) \cite{schulman2017proximal} requires a critic model often as large as the policy itself. To optimize resource use, we adopt group relative policy optimization (GRPO) \cite{shao2024deepseekmath, guo2025deepseek}, which omits the critic by estimating baselines from multiple responses per prompt. Furthermore, training a reward model requires labeled human-preference data, which is costly to collect from radiologists. Following previous RL work in CXR-RG \cite{qin2022reinforced, jing-etal-2019-show}, we utilize the BLEU-4 score between the generated and reference reports as the reward, thereby avoiding the need for a learned reward model. 

While Kullback–Leibler (KL) regularization is the standard safeguard against RL instability, a fixed KL coefficient $\beta$ often creates a ``lose-lose'' scenario: it either stifles productive exploration or fails to prevent destabilizing updates. To address this, recent work has explored adaptive $\beta$ scheduling at the batch level \cite{wu2024beta, schulman2017proximal} or finer-grained adjustments at the instance level \cite{lee2025kl, ICLR2024_c2d82a42}. However, batch-level adjustments often mask harmful individual trajectories by averaging across the batch. Alternatively, instance-level methods either inflate computational overhead with an auxiliary critic \cite{ICLR2024_c2d82a42} or remain strictly dependent on the availability of preference data \cite{lee2025kl}.

To address the above limitations, we propose Response-Weighted Regularization (RER) for fine-grained, response-level control, which is conditioned on both output quality and reference-policy entropy. As a seamless, drop-in replacement for PPO or GRPO objectives, RER requires no auxiliary components or costly preference data. We define a per-response regularization weight $\beta_i$ as a function of the advantage $A_i$ and reference entropy $\mathcal{H}_i^{\mathrm{ref}}$. Specifically, for beneficial exploration ($A_i > 0$), $\beta_i$ is decreased as reference entropy $\mathcal{H}_i^{\mathrm{ref}}$ increases, thereby shielding novel, high-performing behaviors from excessive penalization. Conversely, for harmful deviations ($A_i < 0$), a high $\mathcal{H}_i^{\mathrm{ref}}$ triggers an increase in $\beta_i$ to pull the current policy back toward the reference distribution. By strategically relaxing constraints on productive discovery while tightening them on poor trajectories, RER simultaneously enhances sample efficiency and training stability.

Policy collapse can also occur as the reference policy becomes outdated, causing KL loss to dominate the gradient. While some researchers employ an exponential moving average (EMA) to allow the reference policy to track the current policy \cite{chen2025grpo, rame2024warp}, EMA is inherently vulnerable to reward hacking. If the policy begins to exploit a flawed reward signal, the EMA incorporates this suboptimal behavior into the reference, creating a vicious cycle where future penalties become ineffective at correcting the drift. To resolve these issues, we propose Validation-Anchored Policy Reset (VAPR), a mechanism that resets both the reference policy and current policy to the validation-best checkpoint at fixed intervals. This ensures that every reference update is grounded in empirical validity rather than blind tracking. Unlike the inherent lag in EMA, VAPR's simultaneous update of the reference and current policies nullifies accumulated regularization pressure, providing a fresh, stabilized baseline for continued optimization.

Collectively, RER and VAPR form our Response-Weighted and Validation-Anchored Policy Optimization (REVA-PO) framework. By balancing response-level flexibility with long-term structural stability, REVA-PO enhances the efficiency and quality of exploration while effectively insulating the training process against catastrophic policy collapse.

RL performance is also fundamentally constrained by the quality of the initial policy. A poor initial policy can reinforce stochastic errors rather than meaningful clinical insights. To ensure a robust starting point, we employ a three-stage training pipeline: (1) warm-up, (2) classifier-guided SFT, and (3) RL with REVA-PO. This staged approach ensures that the LVLM begins the RL phase with high-quality exploration and a reliable reference for KL regularization.

Our contributions are summarized as follows:
\begin{itemize}

\item \boldres{Response-Weighted Regularization (RER)}: We propose  RER, a response-level mechanism that dynamically scales $\beta_i$ for each candidate based on its advantage and reference entropy, protecting high-quality discovery while penalizing unstable updates.

\item \boldres{Validation-Anchored Policy Reset (VAPR)}: We propose VAPR to prevent training collapse by periodically resetting the reference and current policies to the best validation checkpoint, clearing accumulated regularization pressure.

\item \boldres{Efficient LVLM-based CXR-RG}: We introduce a three-stage pipeline that leverages a pretrained LVLM to eliminate the need for training a separate image encoder, thereby reducing memory overhead and mitigating overfitting risks on smaller datasets like IU-Xray.

\item \boldres{State-of-the-Art Performance}: Extensive evaluations on MIMIC-CXR and IU-Xray validate the efficacy of REVA-PO. Our method achieves relative BLEU-4 gains of 5.1\% and 3.6\%, respectively, while attaining peak CheXpert F1 (0.506) and RadGraph F1 (0.246) scores. These results confirm superior linguistic fluency and clinical accuracy.
\end{itemize}

\section{Related Work}
\subsection{Chest X-ray Report Generation}
Prior work on CXR-RG spans joint classification, knowledge graphs, RL, and LLMs; each has trade-offs. Adding a disease classification head and training with a joint loss can raise lesion sensitivity \cite{jin2024promptmrg, yang2023radiology, wang2022medical, wang2022automated, wang2022inclusive}, but pushes features toward coarse categories, which hurts descriptive detail. We instead use a frozen classifier to supply disease cues without backpropagating gradients into the generator. Medical knowledge graphs can inject prior knowledge \cite{yang2022knowledge, liu2021exploring, li2023dynamic, huang2023kiut, zhang2020radiology}, but are costly to build and maintain. Causal\mbox{-}intervention methods, such as CMCRL \cite{chen2025cross}, aim to mitigate spurious vision\mbox{-}language correlations and achieve promising results.

RL has been adopted to optimize sequence-level metrics for CXR-RG \cite{qin2022reinforced, wang2021self, wang2022medical}. Most prior approaches rely on self-critical sequence training (SCST) \cite{rennie2017self}, a policy-gradient method with a self-critical baseline for non-differentiable metrics. However, because SCST samples only a single response per prompt, it suffers from high-variance gradients and restricted exploration, which has limited the performance ceiling of RL-based CXR-RG methods. To overcome this, our REVA-PO framework builds upon GRPO by sampling a group of responses per prompt, thereby reducing gradient variance and promoting exploration.

Recently, LLM-based methods have significantly improved report quality \cite{liu2024bootstrapping, jin2025chain, yang2025spatio, wang2025cxpmrg, wang2024hergen, 11142850}. Most of these approaches rely on contrastive learning to align radiographs with LLMs, necessitating large batch sizes and substantial GPU memory. In contrast, we utilize a LVLM \cite{bai2025qwen2} that is pre-aligned on general image–text data. This approach bypasses the need for contrastive pretraining and effectively reduces memory overhead.

\subsection{Reinforcement Learning}
As LVLMs are still emerging, research into LVLM-specific RL remains limited \cite{zhang2025mmrlhf, zhan2025vision}. However, due to their architectural similarities to LLMs, many established RL methods transfer effectively to the multimodal domain \cite{shen2025vlm, zang2025internlm}. In traditional RLHF, PPO \cite{schulman2017proximal} optimizes a clipped objective using a critic and a reward model trained on human preferences. However, this pipeline requires substantial compute and annotation overhead \cite{ouyang2022training, bai2022training, ziegler2019fine}. Direct Preference Optimization (DPO) simplifies this by replacing the reward model with a log-ratio objective between current and reference policies, yet it still necessitates preference pairs \cite{rafailov2023direct}. In contrast, GRPO eliminates the critic and preference pairs by sampling multiple responses per prompt and using group-average rewards as a baseline \cite{shao2024deepseekmath, guo2025deepseek}. Despite these advantages, GRPO relies on a fixed KL $\beta$ that ensures stability at the cost of restricted exploration.

Recent work has explored dynamic $\beta$ adjustment to address this limitation. PPO \cite{schulman2017proximal} adapts $\beta$ based on batch average KL divergence, increasing the penalty when divergence exceeds a target to maintain stability, which can inadvertently throttle learning. $\beta$-DPO \cite{wu2024beta} and $\varepsilon$-DPO \cite{lee2025kl} adapt $\beta$ using signals derived from preference pairs, which remain dependent on paired data. While Zheng et al. \cite{ICLR2024_c2d82a42} use a critic-based classifier to scale $\beta$ for high performing samples, this significantly increases memory costs. Our proposed RER provides fine-grained control for each response without requiring preference data or additional components. Furthermore, a static reference policy allows KL pressure to accumulate, leading to sudden training collapse. While some methods use EMA to update the reference \cite{chen2025grpo, rame2024warp}, blind tracking risks absorbing erroneous behaviors and inherently lags behind the current policy. We instead propose VAPR, which periodically realigns the reference and current policies. This resets the effective KL penalty and ensures long-term training stability.

\section{Method}
\begin{figure*}[!t]
  \centering
   \includegraphics[width=0.95\linewidth]{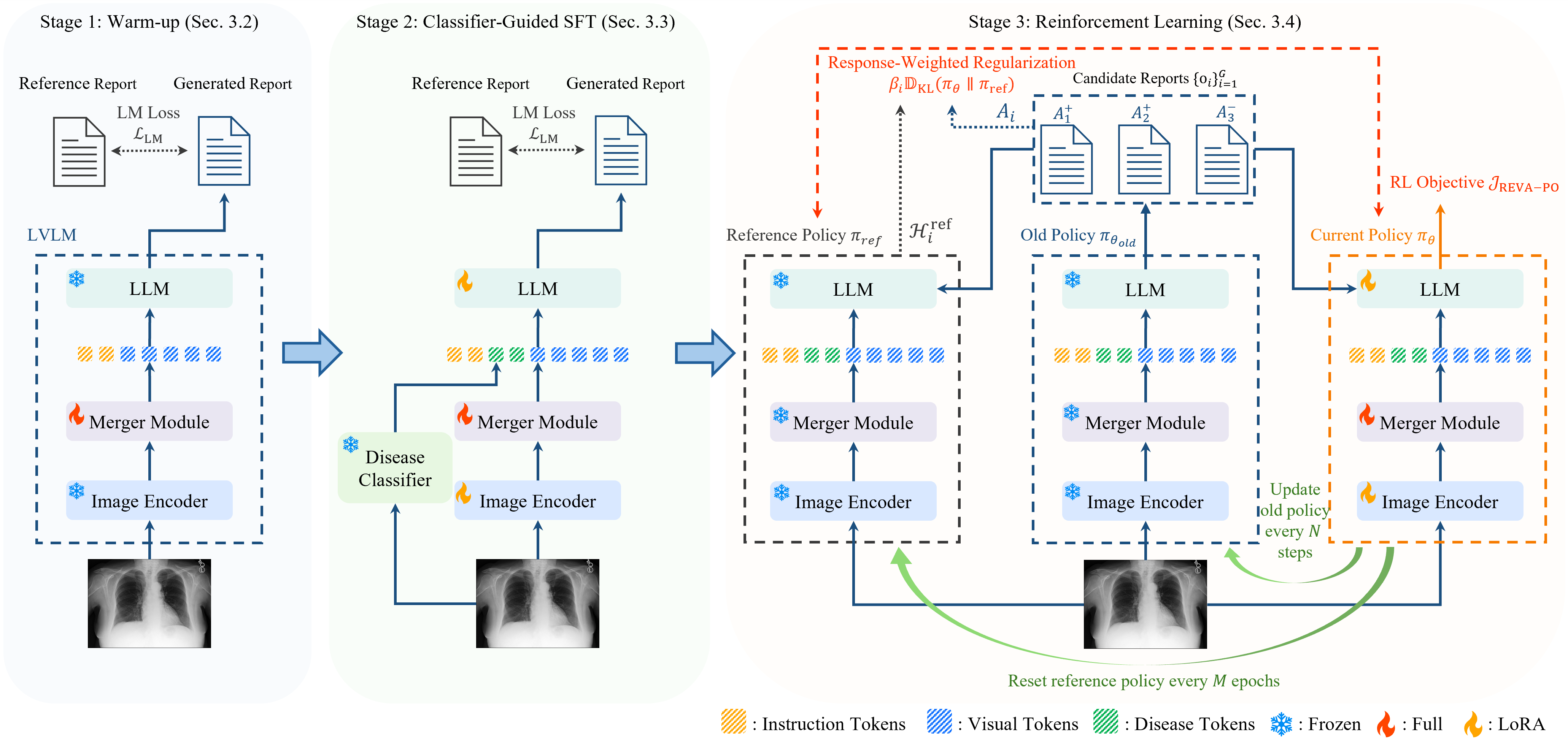}
   \caption{The three\mbox{-}stage training pipeline. \emph{Stage 1: Warm\mbox{-}up.} Freeze the image encoder and LLM; train only the merger with $\mathcal{L}_{\mathrm{LM}}$. \emph{Stage 2: Classifier\mbox{-}guided SFT.} Build prompt $q=[I,V,D]$ by appending predicted labels $D$; freeze the classifier; train the merger and LoRA\mbox{-}tune the encoder and LLM with $\mathcal{L}_{\mathrm{LM}}$. \emph{Stage 3: RL with REVA-PO.} $\pi_{\theta_{\mathrm{old}}}$ samples $G$ reports, computes advantages $A_i$, and maximizes $\mathcal{J}_{\mathrm{REVA-PO}}$ with entropy bonus $\mathcal{H}^{\mathrm{cur}}_{i,t}$ and response-weighted regularization $\beta_i(A_i,\mathcal{H}_i^{\mathrm{ref}})$; reset $\pi_{\mathrm{ref}}$ every $M$ epochs and synchronize $\pi_{\theta_{\mathrm{old}}}$ every $N$ steps. Classifier omitted in Stage~3. The legend indicates instruction, visual, and disease tokens, and the training status of each module (frozen, full, LoRA).}
   \label{fig:pipeline}
\end{figure*}

\boldres{Overview.} We introduce a three-stage pipeline (\cref{fig:pipeline}): (i) a warm-up phase that trains only the merger to adapt the LVLM to radiographs (\cref{sec:warm-up}); (ii) classifier-guided SFT, which injects disease-specific cues (\cref{sec: sft}); and (iii) RL with REVA-PO, which leverages RER to scale KL weights for each response and VAPR to maintain long-term stability while encouraging exploration (\cref{sec:rl}).

\subsection{Warm-up}
\label{sec:warm-up}
An LVLM consists of an image encoder, a merger that maps visual features to the LLM embedding space, and the LLM. Inspired by \cite{liu2024improved,zhu2023minigpt}, we freeze the image encoder and the LLM and train only the merger. This step aligns the scale and distribution of mapped visual features with the radiological domain while keeping the pretrained representations unchanged. With both endpoints fixed, optimization remains stable and the frozen components do not drift.

Given a prompt \(q = [I, V]\) with instruction tokens \(I\) and visual tokens \(V\), and reference tokens \(y_{1:T}\), the language modeling loss is
\begin{equation}
\mathcal{L}_{\mathrm{LM}}(\theta)
= -\frac{1}{T}\sum_{t=1}^{T}\log p_{\theta}\big(y_t \mid y_{<t}, q\big).
\label{lm_loss}
\end{equation}

\subsection{Classifier-Guided Supervised Fine-Tuning}
\label{sec: sft}
After warm-up, the LVLM can produce reports that resemble reference reports in vocabulary and sentence structure, yet its disease-level diagnostic accuracy is still limited. Drawing inspiration from PromptMRG \cite{jin2024promptmrg}, we integrate a pretrained disease classifier to provide explicit clinical cues. Given a CXR, the classifier predicts 14 disease categories following the CheXpert \cite{irvin2019chexpert} taxonomy. We append the predicted positive labels to the prompt to guide the LLM toward lesion-relevant visual tokens.

Concretely, if the classifier predicts “Cardiomegaly” and “Lung Opacity” (denote these as $D$), we form $q=[I,D,V]$, where $I$ are instruction tokens and $V$ are visual tokens. The labels $D$ guide attention to the corresponding anatomical regions and reduce missed findings.

Since the model is warmed up, we increase the number of trainable parameters. The merger is fully trainable, all linear layers of the image encoder and the LLM are fine-tuned with LoRA \cite{hu2022lora}, and the classifier is frozen. Training continues with $\mathcal{L}_{\mathrm{LM}}$ in \cref{lm_loss}.

\subsection{Reinforcement Learning}
\label{sec:rl}
Stage~3 is initialized from the Stage~2 weights and uses the same trainable parameters. Given an input image with prompt $q$ and its reference report $a$, the old policy $\pi_{\theta_{\mathrm{old}}}$ samples a group of $G$ candidate reports $\{o_i\}_{i=1}^{G}$. Each $o_i$ receives a reward $R_i$ given by the BLEU-4 score between $o_i$ and $a$. As illustrated in Supplementary Material A, we use BLEU-4 as a proxy reward for its low variance and strong directional alignment with clinical efficacy metrics. Subsequently, we compute a group-relative standardized advantage
\begin{equation}
  A_i
  \;=\;
  \frac{R_i - \operatorname{mean}\!\big(\{R_j\}_{j=1}^{G}\big)}
       {\operatorname{std}\!\big(\{R_j\}_{j=1}^{G}\big)}.
\end{equation}

\subsubsection{REVA-PO Objective.}
We optimize the policy $\pi_\theta$ using a GRPO-style objective, which incorporates a clipped surrogate loss, the proposed RER mechanism $\beta_i$, and an entropy bonus. The objective is
\begin{equation}
\begin{aligned}
\mathcal{J}_{\mathrm{REVA-PO}}(\theta)
& = \mathbb{E}[(q,a)\sim\mathcal{D},\, \{o_i\}_{i=1}^{G}\sim \pi_{\theta_{\mathrm{old}}}(\cdot\mid q)]
\\[4pt]
& \quad \left[
\frac{1}{\sum_{i=1}^G|o_i|}\sum_{i=1}^{G}\sum_{t=1}^{|o_i|}
\Big(\ell_{i,t}(\theta)-\beta_i\, \mathbb{D}_{\mathrm{KL}}\!\left(\pi_{\theta}\,\|\,\pi_{\mathrm{ref}}\right)
\Big)
\right],
\end{aligned}
\label{eq:REVA-PO}
\end{equation}
where the clipped term is
\begin{equation}
    \ell_{i,t}(\theta) = \min\!\Big(r_{i,t}(\theta)\,A_i,\;
\operatorname{clip}\!\left(r_{i,t}(\theta),\,1-\epsilon_{\text{low}},\,1+\epsilon_{\text{high}}\right)A_i\Big) ,
\end{equation}
with importance ratio
\begin{equation}
  r_{i,t}(\theta)
  \;=\;
  \frac{\pi_{\theta}\!\left(o_{i,t}\mid q, o_{i,<t}\right)}
       {\pi_{\theta_{\mathrm{old}}}\!\left(o_{i,t}\mid q, o_{i,<t}\right)}.
  \label{eq:ratio}
\end{equation}
The Kullback-Leibler (KL) divergence uses the unbiased estimator
\begin{equation}
\begin{aligned}
\mathbb{D}_{\mathrm{KL}}\!\left(\pi_{\theta}\,\|\,\pi_{\mathrm{ref}}\right) & = \frac{\pi_{\mathrm{ref}}\!\left(o_{i,t}\mid q, o_{i,<t}\right)}
{\pi_{\theta}\!\left(o_{i,t}\mid q, o_{i,<t}\right)} \\
&-\log \frac{\pi_{\mathrm{ref}}\!\left(o_{i,t}\mid q, o_{i,<t}\right)}
{\pi_{\theta}\!\left(o_{i,t}\mid q, o_{i,<t}\right)} -1.
\end{aligned}
\end{equation}

As suggested by \cite{yu2025dapo}, we use the token-level average $\frac{1}{\sum_{i}|o_i|}\sum_{i}\sum_{t}(\cdot)$ in \cref{eq:REVA-PO} to give each token equal weight and avoid downweighting long responses. Maximizing $\mathcal{J}_{\mathrm{REVA-PO}}$ increases the probability of high-advantage responses and decreases that of low-advantage responses.

\subsubsection{Response-Weighted Regularization.}
KL regularization stabilizes learning by penalizing deviation from the reference policy $\pi_{\mathrm{ref}}$. However, a fixed weight $\beta$ can over-penalize beneficial exploration and under-penalize harmful drifts. We therefore introduce RER, which assigns a response-level coefficient $\beta_i$ to each candidate $o_i$ based on its advantage $A_i$ and the reference-policy entropy along the sequence. The sequence-level reference-policy entropy is defined as
\begin{equation}
  \mathcal{H}^{\mathrm{ref}}_i
  =-\frac{1}{|o_i|}\sum_{t=1}^{|o_i|}\sum_{v\in\mathcal{V}} \pi_{\mathrm{ref}}(v \mid q, o_{i,<t})\,\log \pi_{\mathrm{ref}}(v \mid q, o_{i,<t}),
  \label{eq:Href}
\end{equation}
where $\mathcal{V}$ is the token vocabulary. Using the token-normalized entropy makes $\mathcal{H}^{\mathrm{ref}}_i$ insensitive to response length. We normalize this entropy within the group to obtain
\begin{equation}
  \hat{\mathcal{H}}^{\mathrm{ref}}_i
  \;=\;
  \frac{\mathcal{H}^{\mathrm{ref}}_i - \operatorname{min}\!\big(\{\mathcal{H}^{\mathrm{ref}}_j\}_{j=1}^{G}\big)}
       {\operatorname{max}\!\big(\{\mathcal{H}^{\mathrm{ref}}_j\}_{j=1}^{G}\big) - \operatorname{min}\!\big(\{\mathcal{H}^{\mathrm{ref}}_j\}_{j=1}^{G}\big)} \;\in\;[0,1].
  \label{eq:group_adv}
\end{equation}
We then set $\beta_i$ as a function of $A_i$ and $\hat{\mathcal{H}}^{\mathrm{ref}}_i$
\begin{equation}
\beta_i=
\begin{cases}
\beta_{\mathrm{min}}+(\beta_{\mathrm{base}}-\beta_{\mathrm{min}})\,(1-\hat{\mathcal{H}}^{\mathrm{ref}}_i), & A_i > 0,\\[1pt]
\beta_{\mathrm{base}}+(\beta_{\mathrm{max}}-\beta_{\mathrm{base}})\,\hat{\mathcal{H}}^{\mathrm{ref}}_i, & A_i < 0,\\[1pt]
\beta_{\mathrm{base}}, & A_i = 0,
\end{cases}
\label{eq:beta_piecewise}
\end{equation}
with $\beta_{\mathrm{min}}\le \beta_{\mathrm{base}}\le \beta_{\mathrm{max}}$.

(i) For $A_i>0$ (better-than-mean response), a larger $\hat{\mathcal{H}}^{\mathrm{ref}}_i$ indicates the reference policy is uncertain about producing this good response. We therefore reduce $\beta_i$ toward $\beta_{\mathrm{min}}$ to avoid constraining promising updates. 

(ii) For $A_i<0$ (worse-than-mean response), a larger $\hat{\mathcal{H}}^{\mathrm{ref}}_i$ implies the reference policy is less likely to generate such a poor response. We therefore increase $\beta_i$ toward $\beta_{\mathrm{max}}$, which strengthens the pull that drives $\pi_{\theta}$ back toward $\pi_{\mathrm{ref}}$.

\begin{figure*}[!tb]
  \centering
  \begin{subfigure}[t]{0.205\textwidth}
    \centering
    \includegraphics[width=\linewidth]{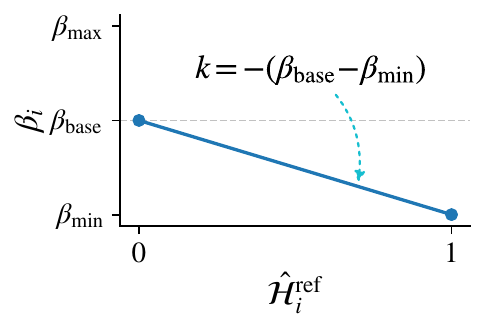}
    \subcaption{$A_i>0$}
    \label{fig:beta_h_Apos}
  \end{subfigure}\hfill
  \begin{subfigure}[t]{0.205\textwidth}
    \centering
    \includegraphics[width=\linewidth]{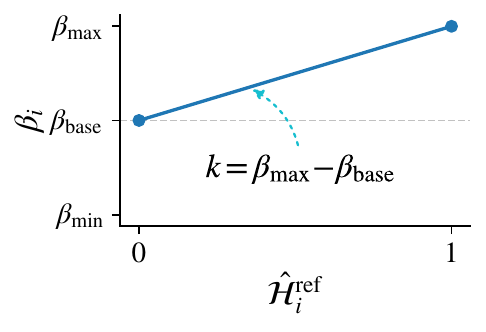}
    \subcaption{$A_i<0$}
    \label{fig:beta_h_Aneg}
  \end{subfigure}\hfill
  \subcaptionbox{Fixed Reference\label{fig:fixed_reference}}{%
    \includegraphics[height=0.155\textwidth]{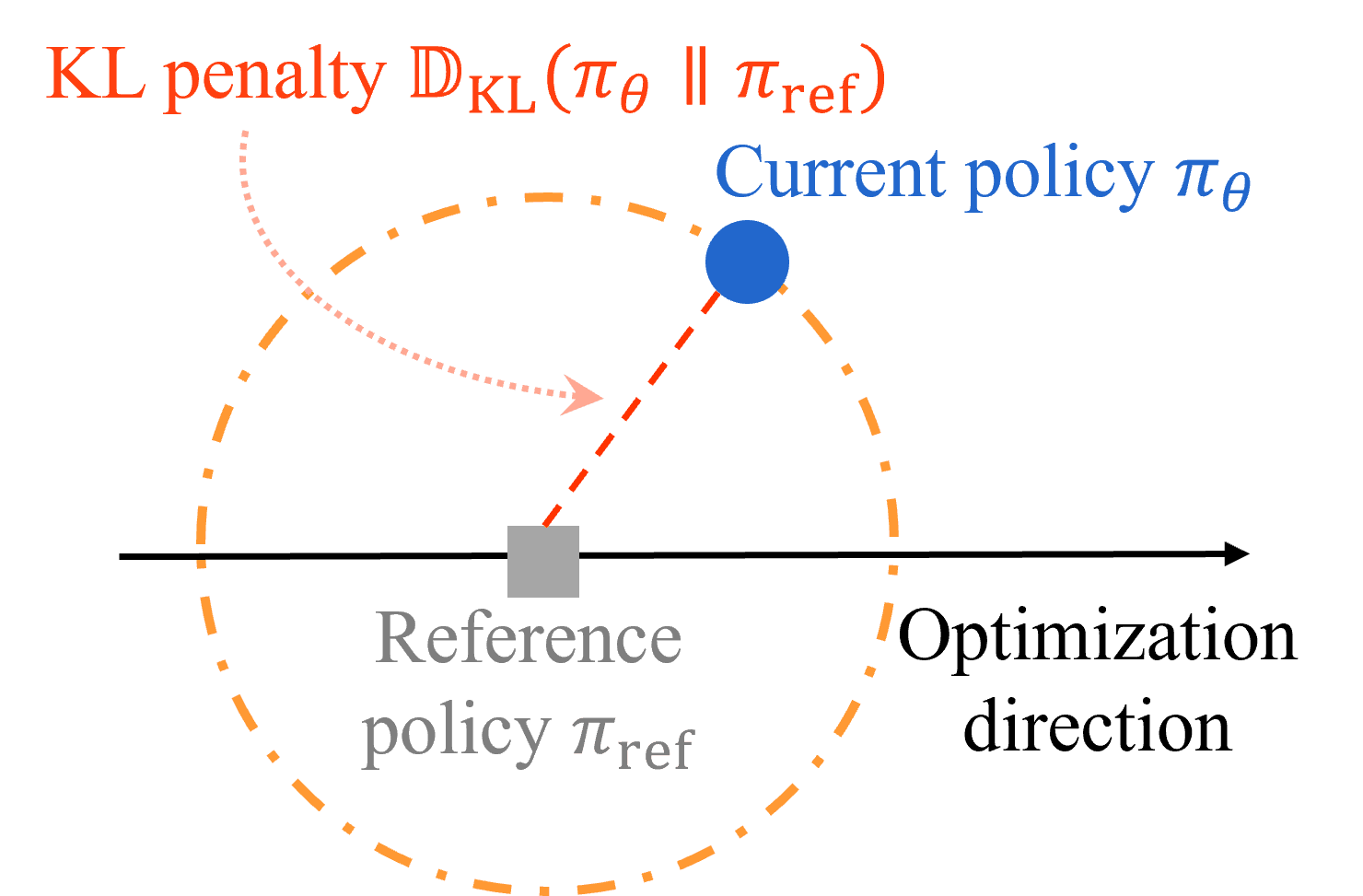}%
  }\hfill
  \subcaptionbox{VAPR\label{fig:periodic_reference}}{%
    \includegraphics[height=0.155\textwidth]{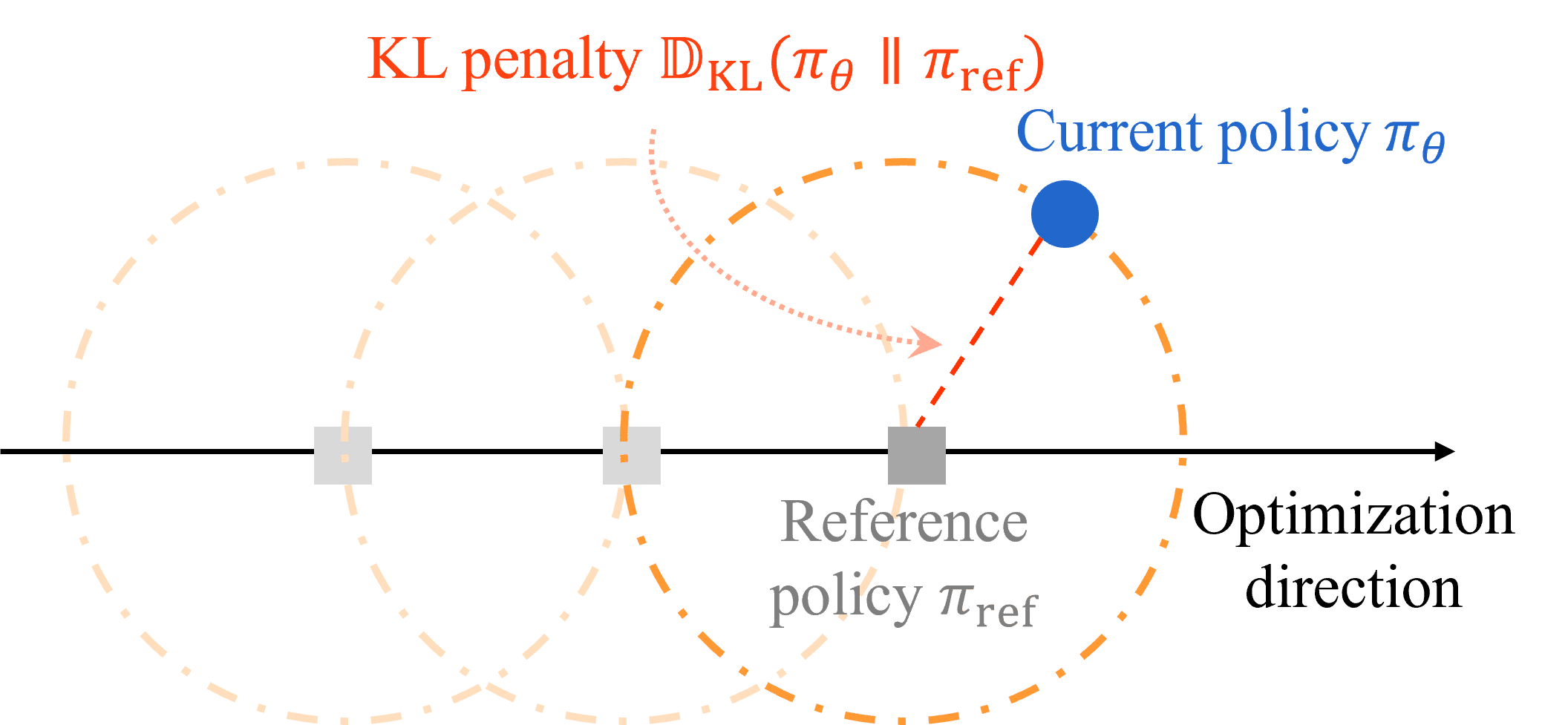}%
  }

  \caption{
  (Left) Mapping from the normalized reference-policy entropy $\hat{\mathcal{H}}^{\mathrm{ref}}_i$ to the per-response KL weight $\beta_i$.
  For advantage $A_i>0$, the line has slope $k=-(\beta_{\mathrm{base}}-\beta_{\min})$; for $A_i<0$, the line has slope $k=\beta_{\max}-\beta_{\mathrm{base}}$.
  (Right) Fixed Reference vs.\ Validation-Anchored Policy Reset (VAPR).
  The reference policy $\pi_{\mathrm{ref}}$ is the anchor (gray), the current policy $\pi_\theta$ is the ball (blue), and the KL penalty $\mathbb{D}_{\mathrm{KL}}(\pi_\theta \,\|\, \pi_{\mathrm{ref}})$ is the elastic string.
  With fixed $\pi_{\mathrm{ref}}$, the KL term keeps exploration local; VAPR resets $\pi_{\mathrm{ref}}$ and $\pi_\theta$ to the validation\mbox{-}best checkpoint, effectively releasing the regularization pressure to expand the reachable region.
  }
  \label{fig:beta_h_and_VAPR_one_row}
\end{figure*}

The mapping $(A_i,\hat{\mathcal{H}}^{\mathrm{ref}}_i)\mapsto \beta_i$ is bounded and piecewise monotonic with respect to $\hat{\mathcal{H}}^{\mathrm{ref}}_i$ (\cref{fig:beta_h_Apos,fig:beta_h_Aneg}), preserving the numerical stability of $\beta_i$ throughout the training process.

\subsubsection{Validation-Anchored Policy Reset.}
\label{sec: periodic_update}
Although RER relaxes penalties on promising exploratory actions, exploration remains fundamentally constrained by $\mathbb{D}_{\mathrm{KL}}\!\left(\pi_{\theta}\,\|\,\pi_{\mathrm{ref}}\right)$ when $\pi_{\mathrm{ref}}$ is fixed. As $\pi_\theta$ drifts from $\pi_{\mathrm{ref}}$, the growing KL term pulls the policy back, restricting exploration to a static trust region around the initial reference.

To overcome this, we introduce VAPR, which periodically expands the explorable space while maintaining stability. After each epoch, we evaluate $\pi_\theta$ on the validation set and save the checkpoint $\theta_{\mathrm{best}}$ that yields the highest validation score $s_\text{val}$: 
\begin{equation}
\label{eq:s_val}
s_{\mathrm{val}} = \frac{1}{|\mathcal{M}| + 1} \left( \sum_{m \in \mathcal{M}} m + \lambda \cdot \mathrm{RG} \right),
\end{equation}
where $\mathcal{M}$ is the set of natural language generation metrics, $\mathrm{RG}$ denotes the RadGraph F1 \cite{delbrouck2024radgraph} as a proxy for clinical factual consistency, and $\lambda$ is a balancing coefficient.

Every $M$ epochs, we reset $\pi_{\mathrm{ref}}$ and synchronize the $\pi_{\theta_{\mathrm{old}}}$ and $\pi_\theta$ to $\theta_{\mathrm{best}}$. This reset enforces $\mathbb{D}_{\mathrm{KL}}\!\left(\pi_{\theta}\,\|\,\pi_{\mathrm{ref}}\right)=0$, allowing optimization to proceed from the new anchor without accumulated regularization pressure. Synchronizing $\pi_{\theta_{\mathrm{old}}}$ simultaneously mitigates off-policy sampling mismatch. \cref{alg:REVA-PO_clean} outlines the full procedure.

\Cref{fig:fixed_reference,fig:periodic_reference} illustrate this intuition: treating $\pi_{\mathrm{ref}}$ as an anchor, $\pi_\theta$ as a movable ball, and $\mathbb{D}_{\mathrm{KL}}\!\left(\pi_{\theta}\,\|\,\pi_{\mathrm{ref}}\right)$ as an elastic string. A fixed anchor tightens the string as the ball moves away, confining exploration locally. By contrast, VAPR relocates the anchor to the best-performing region discovered so far. The ball then explores around this updated position, effectively extending the reachable optimization path without straying into unstable regions.

\begin{algorithm}[!tb]
  \small
  \caption{REVA-PO: Response-Weighted and Validation-Anchored Policy Optimization}
  \label{alg:REVA-PO_clean}
  \begin{algorithmic}[1]
    \Require policy $\pi_{\theta}$;
            dataset $\mathcal{D}$; group size $G$; learning rate $\eta$; sampling synchronization period $N$ (steps); reference reset period $M$ (epochs); total epochs $E$; steps per epoch $S$
    \State $\pi_{\mathrm{ref}} \gets \pi_{\theta}$,\quad $\pi_{\theta_{\mathrm{old}}} \gets \pi_{\theta}$,\quad $\theta_{\mathrm{best}} \gets \theta$
    \For{$\mathrm{epoch}=1$ \boldres{to} $E$}
      \For{$\mathrm{step}=1$ \boldres{to} $S$}
          \State sample $(q,a) \sim \mathcal{D}$
          \State sample $G$ candidates $\{o_i\}_{i=1}^{G} \sim \pi_{\theta_{\mathrm{old}}}(\cdot \mid q)$
          \State compute rewards $\{R_i\}_{i=1}^{G}$ for the sampled $\{o_i\}$
        \State compute $\mathcal{J}_{\mathrm{REVA-PO}}(\theta;\,\{o_i\},\pi_{\mathrm{ref}})$
        \Comment{\cref{eq:REVA-PO}}
        \State $\theta \gets \theta + \eta\,\nabla_{\theta}\mathcal{J}_{\mathrm{REVA-PO}}(\theta;\,\{o_i\},\pi_{\mathrm{ref}})$
        \If{$\mathrm{step} \bmod N = 0$}
          \State $\pi_{\theta_{\mathrm{old}}} \gets \pi_{\theta}$ \Comment{synchronize old (sampling) policy}
        \EndIf
      \EndFor
      \State evaluate $\pi_{\theta}$ on validation set to obtain $s_{\mathrm{val}}$
      \If{$s_{\mathrm{val}}$ is the new best}
        \State $\theta_{\mathrm{best}} \gets \theta$
      \EndIf
      \If{$\mathrm{epoch} \bmod M = 0$}
      \Comment{\cref{sec: periodic_update}}
        \State $\pi_{\mathrm{ref}} \gets \pi_{\theta_{\mathrm{best}}}$ 
        \State $\pi_\theta \gets \pi_{\theta_{\mathrm{best}}}$ 
        \State $\pi_{\theta_{\mathrm{old}}} \gets \pi_\theta$ \Comment{reset and synchronize all policies}
      \EndIf
    \EndFor
    \Ensure optimized policy $\pi_{\theta}$
  \end{algorithmic}
\end{algorithm}

\section{Experiments}
\subsection{Datasets and Metrics}
We evaluate on two widely used CXR-RG benchmarks: MIMIC-CXR \cite{johnson2019mimic} and IU-Xray \cite{demner2015preparing}. Following prior work \cite{wang2025cxpmrg, liu2024bootstrapping,shen2024automatic}, we report natural language generation (NLG) and clinical efficacy (CE) metrics.

\boldres{MIMIC-CXR.} The MIMIC-CXR dataset, collected at Beth Israel Deaconess Medical Center, contains 377{,}110 chest radiographs and 227{,}827 radiology reports. We use the official train/validation/test split for fair comparison with prior work.

\boldres{IU-Xray.} The IU-Xray dataset, developed by Indiana University, contains 7{,}470 chest radiographs and 3{,}955 reports. Following prior work \cite{chen-acl-2021-r2gencmn, li2023dynamic, wang2023metransformer}, we use the established 7:1:2 train/validation/test split.

\boldres{Metrics.} We assess the descriptive accuracy of generated reports with NLG metrics: BLEU \cite{papineni2002bleu}, ROUGE-L \cite{lin2004rouge}, and METEOR \cite{banerjee2005meteor}. For CE metrics, we apply the CheXpert\footnote{https://github.com/stanfordmlgroup/chexpert-labeler} labeler to extract 14 disease categories from generated and reference reports. We binarize labels (positive as 1; negative, uncertain, and not mentioned as 0) and compute precision, recall, and F1. We also use RadGraph F1 \cite{delbrouck2024radgraph} to evaluate the consistency of clinical entities (e.g., anatomical structures and lesion types) and their relationships in the generated reports.

\subsection{Implementation Details}
In the preliminary study, we compared two popular open-source LVLMs: Qwen2.5-VL-7B-Instruct \cite{bai2025qwen2} and Llama-3.2-11B-Vision-Instruct \cite{grattafiori2024llama}. While both models achieved similar performance in Stage 1 and Stage 2, Llama required significantly more memory and was only stable using bfloat16 (BF16); float16 (FP16) training resulted in exploding gradients. In contrast, Qwen consumed less memory and maintained stability with both BF16 and FP16. Consequently, we adopted Qwen2.5-VL-7B-Instruct as our base model. All experiments were conducted on 8 NVIDIA Tesla V100 GPUs using FP16. The disease classifier, fine-tuned from \cite{xiao2023delving}, remained frozen during Stage 2 and Stage 3. Full implementation details and hyperparameters are available in the Supplementary Material.

\subsection{Quantitative Results}

\subsubsection{NLG Metrics.} \Cref{tab:main_results} compares REVA-PO against representative and state-of-the-art methods on the MIMIC-CXR and IU-Xray datasets. Details of these methods are provided in Supplementary Material H. REVA-PO ranks first in five of six NLG metrics on IU-Xray and four of six on MIMIC-CXR. Specifically, on IU-Xray, our method surpasses strong baselines such as CMCRL \cite{chen2025cross} and AM-MRG \cite{11142850} across BLEU-1 through BLEU-4 and ROUGE-L. Notably, BLEU-3 and BLEU-4 increase by $+3.2\% $ (0.253 $\rightarrow$ 0.261) and $+3.6\%$ (0.192 $\rightarrow$ 0.199), respectively. We observe similar trends on MIMIC-CXR, where REVA-PO outperforms the previous best, AM-MRG \cite{11142850}, with relative gains of $+4.3\%$ for BLEU-3 (0.187 $\rightarrow$ 0.195) and $+5.1\%$ for BLEU-4 (0.136 $\rightarrow$ 0.143).

Among methods evaluated on both datasets, most LLM-based approaches (MambaXray-VL~\cite{wang2025cxpmrg}, STREAM~\cite{yang2025spatio}, CoD~\cite{jin2025chain}, B-LLM~\cite{liu2024bootstrapping}) exhibit strong performance on the large-scale MIMIC-CXR but fail to surpass the non-LLM method CMCRL \cite{chen2025cross} on the smaller IU-Xray. This performance drop likely stems from a susceptibility to overfitting when training specialized image encoders on limited medical data. In contrast, our approach leverages an LVLM without the need for additional image-encoder pre-training, thereby mitigating overfitting risks. REVA-PO consistently outperforms both LLM and non-LLM baselines across both datasets, demonstrating superior generalization across varying data scales and complexities.

\subsubsection{CE Metrics.} 
\Cref{tab:mimic_ce} presents the CE metrics on MIMIC-CXR, where REVA-PO establishes new benchmarks in Chexpert precision (0.569), Chexpert F1 (0.506), and RadGraph F1 (0.246). Regarding diagnostic accuracy, our method surpasses AM-MRG \cite{11142850} by $4.5\%$ in CheXpert F1 (0.484 $\rightarrow$ 0.506). Moreover, REVA-PO outperforms the previous state-of-the-art, STREAM \cite{yang2025spatio}, by $12.8\%$ in RadGraph F1 (0.218 $\rightarrow$ 0.246). This substantial margin underscores that our approach generates reports with superior clinical factual consistency with reference reports.
\begin{table}[!t]
  \centering
\scriptsize
\captionsetup{justification=raggedright,singlelinecheck=false}
  \caption{Comparisons of NLG metrics between our REVA-PO and state-of-the-art methods on IU-Xray (top) and MIMIC-CXR (bottom) using BLEU-1 to BLEU-4 (B-1--4), ROUGE-L (R-L), and METEOR (MTR). Best scores are in bold.}
  \label{tab:main_results}
  \setlength{\tabcolsep}{4pt}
  \renewcommand{\arraystretch}{1.10}

  \begin{tabular}{>{\centering\arraybackslash}m{1.2cm}|>{\centering\arraybackslash}m{1.13cm}|>{\raggedright\arraybackslash}p{2.45cm}|c|*6{S[table-format=1.3]}}
  \toprule
    \boldres{Datasets} & \boldres{Types} & \boldres{Methods} & \boldres{Years} &
    \boldres{B-1} & \boldres{B-2} & \boldres{B-3} & \boldres{B-4} & \boldres{R-L} & \boldres{MTR} \\
    \midrule
    
    \multirow{13}{*}{\boldres{IU}}
      &  \multirow{7}{*}{non-LLM}                        & Clinical-BERT \cite{yan2022clinical}    & 2022 & 0.495 & 0.330 & 0.231 & 0.170 & 0.376 & 0.209 \\
      &                          & CMMRL \cite{qin2022reinforced}                & 2022 & 0.494 & 0.321 & 0.235 & 0.181 & 0.384 & 0.201 \\
      &                          & DCL \cite{li2023dynamic}                & 2023 & \text{--} & \text{--} & \text{--} & 0.163 & 0.383 & 0.193 \\
      &                          & METransformer \cite{wang2023metransformer} & 2023 & 0.483 & 0.322 & 0.228 & 0.172 & 0.380 & 0.192 \\
      &                          & PromptMRG \cite{jin2024promptmrg}       & 2024 & 0.401 & \text{--} & \text{--} & 0.098 & 0.281 & 0.160 \\
      &                          & MA \cite{shen2024automatic}             & 2024 & 0.501 & 0.328 & 0.230 & 0.170 & 0.386 & 0.213 \\
      &                          & $\text{CMCRL}_{3}$ \cite{chen2025cross} & 2025 & 0.505 & 0.334 & 0.245 & 0.190 & 0.394 & 0.210 \\
    \cmidrule{2-10}
      & \multirow{5}{*}{LLM} &B-LLM \cite{liu2024bootstrapping} & 2024 & 0.499 & 0.323 & 0.238 & 0.184 & 0.390 & 0.208 \\
      &                            & CoD \cite{jin2025chain}                & 2025 & 0.403 & \text{--} & \text{--} & 0.091 & 0.288 & \text{--} \\
      &                            & STREAM \cite{yang2025spatio}   & 2025 & 0.499 & 0.333 & 0.238 & 0.178 & 0.377 & 0.213 \\
      &                            & MambaXray-VL \cite{wang2025cxpmrg}   & 2025 & 0.491 & 0.330 & 0.241 & 0.185 & 0.371 & 0.216 \\
      &                            & AM-MRG \cite{11142850}   & 2026 & 0.489 & 0.339 & 0.253 & 0.192 & 0.384 & \boldres{0.225} \\
       \cmidrule{2-10}
      & \multirow{1}{*}{LVLM} 
      & REVA-PO (ours)                            &      & \boldres{0.517} & \boldres{0.356} & \boldres{0.261} & \boldres{0.199} & \boldres{0.406} & 0.216 \\
    \midrule

    \multirow{17}{*}{\boldres{MIMIC}}
      & \multirow{9}{*}{non-LLM}                      & Clinical-BERT \cite{yan2022clinical}    & 2022 & 0.383 & 0.230 & 0.151 & 0.106 & 0.275 & 0.144 \\
      &                          & CMMRL \cite{qin2022reinforced}                & 2022 & 0.381 & 0.232 & 0.155 & 0.109 & 0.287 & 0.151 \\
      &                          & DCL \cite{li2023dynamic}                & 2023 & \text{--} & \text{--} & \text{--} & 0.109 & 0.284 & 0.150 \\
      &                          & METransformer \cite{wang2023metransformer} & 2023 & 0.386 & 0.250 & 0.169 & 0.124 & 0.291 & 0.152 \\
      &                          & NADM \cite{zhao2023normal}                & 2023 & 0.402 & 0.258 & 0.179 & 0.130 & 0.289 & 0.155 \\
      &                          & PromptMRG \cite{jin2024promptmrg}       & 2024 & 0.398 & \text{--} & \text{--} & 0.112 & 0.268 & 0.157 \\
      &                          & MA \cite{shen2024automatic}             & 2024 & 0.396 & 0.244 & 0.162 & 0.115 & 0.274 & 0.151 \\
      &                          & RRG-DPO \cite{liu2025rrg}             & 2025 & \text{--} & \text{--} & \text{--} & 0.112 & 0.286 & 0.166 \\
      &                          & $\text{CMCRL}_{6}$ \cite{chen2025cross} & 2025 & 0.400 & 0.245 & 0.165 & 0.119 & 0.280 & 0.150 \\
    \cmidrule{2-10}
      & \multirow{7}{*}{LLM} & RGRG \cite{tanida2023interactive} & 2023 & 0.373 & 0.249 & 0.175 & 0.126 & 0.264 & 0.168 \\
      &                          & HERGen \cite{wang2024hergen}            & 2024 & 0.395 & 0.248 & 0.169 & 0.122 & 0.285 & 0.156 \\
      & & B-LLM \cite{liu2024bootstrapping} & 2024 & 0.402 & 0.262 & 0.180 & 0.128 & 0.291 & \boldres{0.175} \\
      &                            & CoD \cite{jin2025chain}                & 2025 & 0.412 & \text{--} & \text{--} & 0.129 & 0.286 & \text{--} \\
      &                            & STREAM \cite{yang2025spatio}   & 2025 & 0.420 & 0.267 & 0.184 & 0.133 & 0.291 & 0.164 \\
      &                            & MambaXray-VL \cite{wang2025cxpmrg}   & 2025 & 0.422 & 0.268 & 0.184 & 0.133 & 0.289 & 0.167 \\
      &                            & AM-MRG \cite{11142850}   & 2026 & \boldres{0.426} & 0.271 & 0.187 & 0.136 & 0.291 & 0.174 \\
       \cmidrule{2-10}
      & \multirow{1}{*}{LVLM} 
      & REVA-PO (ours)                            &      & 0.424 & \boldres{0.280} & \boldres{0.195} & \boldres{0.143} & \boldres{0.293} & 0.170 \\
    \bottomrule
  \end{tabular}
\end{table}

\begin{table}[!tb]
  \centering
    \scriptsize
    \captionsetup{justification=raggedright,singlelinecheck=false}
  \caption{Comparisons of the CE metrics on MIMIC-CXR using CheXpert precision (P), CheXpert recall (R), CheXpert F1 (F1), and RadGraph F1 (RG). Best scores are in bold.}
  \label{tab:mimic_ce}
  \setlength{\tabcolsep}{7pt}
  \renewcommand{\arraystretch}{1.10}

  \begin{tabular}{>{\raggedright\arraybackslash}p{2.6cm}|*4{S[table-format=1.3]}}
  \toprule
    \boldres{Methods} & \boldres{P} & \boldres{R} & \boldres{F1} & \boldres{RG} \\
    \midrule
    CMMRL \cite{qin2022reinforced}   & 0.342 & 0.294 & 0.292 & \text{--}\\
    METransformer \cite{wang2023metransformer}   & 0.346 & 0.309 & 0.311 & \text{--}\\
    MambaXray-VL \cite{wang2025cxpmrg}         & 0.371 & 0.321 & 0.340 & \text{--}\\
    HERGen \cite{wang2024hergen}                 & 0.415 & 0.301 & 0.371 & \text{--}\\
    DCL \cite{li2023dynamic}                     & 0.471 & 0.352 & 0.373 & \text{--}\\
    MA \cite{shen2024automatic}                 & 0.411 & 0.398 & 0.389 & \text{--}\\
    $\text{CMCRL}_{6}$ \cite{chen2025cross}      & 0.489 & 0.340 & 0.401 & \text{--}\\
    Clinical-BERT \cite{yan2022clinical}         & 0.387 & 0.435 & 0.415 & \text{--}\\
    NADM \cite{zhao2023normal}                  & 0.417 & 0.413 & 0.415 & 0.209\\
    RRG-DPO \cite{liu2025rrg}                  & 0.462 & 0.487 & 0.443 & 0.207\\
    RGRG \cite{tanida2023interactive}                 & 0.461 & 0.475 & 0.447 & \text{--}\\
    STREAM \cite{yang2025spatio}         & 0.531 & 0.407 & 0.460 & 0.218\\
    B-LLM \cite{liu2024bootstrapping} & 0.465 & 0.482 & 0.473 & \text{--}\\
    PromptMRG \cite{jin2024promptmrg} & 0.501 & 0.509 & 0.476 & \text{--} \\
    CoD \cite{jin2025chain} & 0.487 & \boldres{0.521} & 0.479 & \text{--}\\
    AM-MRG \cite{11142850} & 0.555 & 0.429 & 0.484 & \text{--}\\
    \midrule    
    REVA-PO (ours)                                  & \boldres{0.569} & 0.455 & \boldres{0.506} & \boldres{0.246}\\
    \bottomrule
  \end{tabular}
\end{table}

\subsection{Ablation Study}
We conduct ablation studies to quantify the individual contributions of our three-stage pipeline and the modifications introduced in the RL stage (RER and VAPR). Additional stability analyses are available in the Supplementary Material.

\boldres{Effect of the Three-Stage Pipeline.} From \cref{tab:ablation_stages}, Stage 2 improves CE metrics by $13.5\%$ compared to Stage 1, a gain primarily attributable to the integration of disease cues from the classifier. Notably, recall rises from 0.353 to 0.431 ($+22.1\%$), suggesting that the classifier effectively mitigates the issue of missed findings.

Stage 3 (RL) further enhances performance, yielding a $15.6\%$ increase in NLG metrics and a $22.1\%$ improvement in CE metrics over the Stage 1 baseline. These results indicate that RL can simultaneously improve linguistic overlap with reference reports and diagnostic accuracy.

The final three rows of \cref{tab:ablation_stages} confirm that both Stage 1 and Stage 2 are essential for reaching peak performance.

\begin{figure*}[t!]
  \captionsetup{justification=raggedright,singlelinecheck=false}
  \centering

  \begin{subfigure}{0.42\textwidth}
    \centering
    \includegraphics[width=\linewidth]{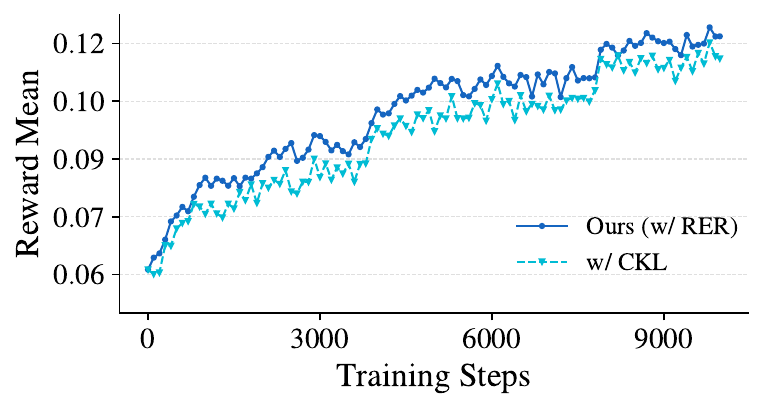}
    \caption{w/ RER vs. w/ CKL\label{fig: ckl}}
    \label{fig:stage3_a}
  \end{subfigure}\hfil
  \begin{subfigure}{0.42\textwidth}
    \centering
    \includegraphics[width=\linewidth]{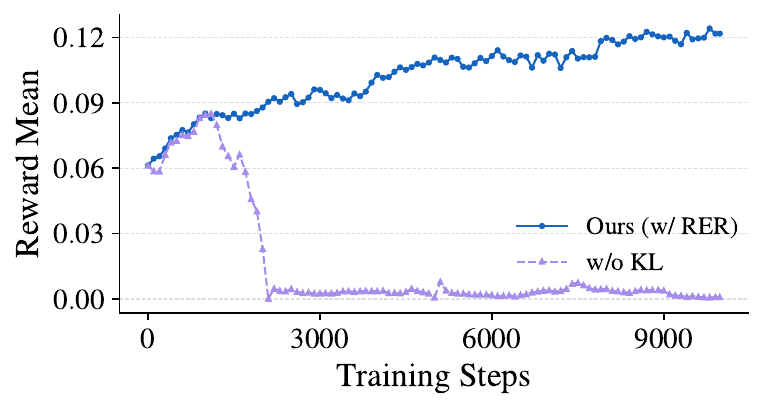}
    \caption{w/ RER vs. w/o KL\label{fig: wokl}}
    \label{fig:stage3_b}
  \end{subfigure}
  
  \begin{subfigure}{0.42\textwidth}
    \centering
    \includegraphics[width=\linewidth]{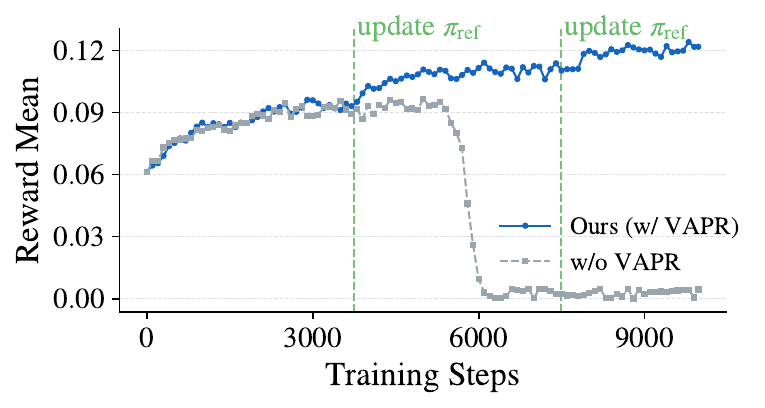}
    \caption{w/ VAPR vs. w/o VAPR\label{fig: VAPR}}
  \end{subfigure}\hfil
  \begin{subfigure}{0.42\textwidth}
    \centering
    \includegraphics[width=\linewidth]{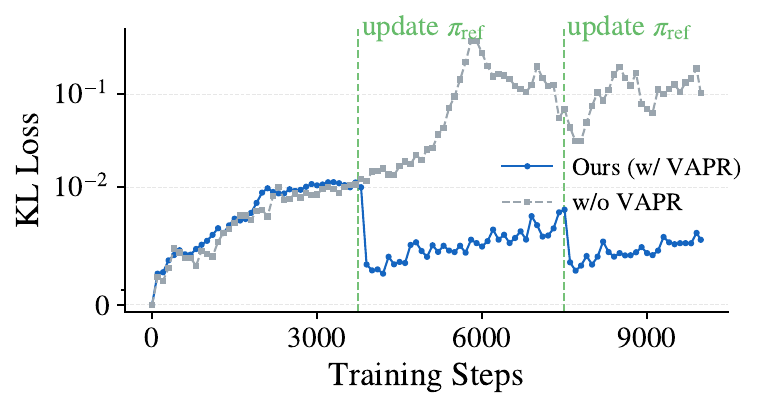}
    \caption{w/ VAPR vs. w/o VAPR\label{fig: VAPR_kl}}
  \end{subfigure}
  \caption{Ablation study of RL-stage modifications on MIMIC-CXR. Mean reward of responses sampled from $\pi_{\theta_{\mathrm{old}}}$ \vs training steps. (a) Response-Weighted Regularization (RER) outperforms constant KL (CKL). (b) Omitting the KL term causes early collapse. (c) Validation-Anchored Policy Reset (VAPR) prevents collapse and boosts rewards; green lines mark $\pi_{\mathrm{ref}}$ updates. (d) KL loss \vs training steps, plotted on a symmetric logarithmic scale to accommodate the wide value range in later training.}

  \label{fig:stage3_row_three}
\end{figure*}

\boldres{Effect of Response-Weighted Regularization.} We monitor the average reward of responses sampled from $\pi_{\theta_{\mathrm{old}}}$ during training. As illustrated in \cref{fig: ckl}, the reward curve for RER ($\beta=\beta_i$) consistently remains above the constant-KL baseline (CKL; $\beta=\beta_{\mathrm{base}}$). This gap suggests that response-level adaptive scaling improves sample efficiency, leading to faster convergence and a higher reward plateau. These gains are especially important in data-constrained settings such as medical report generation. This observation is further corroborated by the empirical results in \cref{tab:ablation_settings}, where the configuration using RER (row~1) achieves higher final scores than the CKL counterpart (row~3).

Although some recent studies advocate removing the KL regularizer to encourage exploration \cite{hu2025open,yu2025dapo,cui2025process,yan2025learning}, we observe training collapse in its absence. \Cref{fig: wokl} shows that removing the KL term drives sampling rewards toward zero early in training, likely due to the policy drifting into low-density regions of the LVLM's prior. This confirms that KL regularization remains indispensable for stable RL in the CXR-RG task.

\boldres{Effect of Validation-Anchored Policy Reset.} As seen in \cref{fig: VAPR}, omitting VAPR leads to training collapse at approximately 6,000 steps. Upon the first reference update, a clear performance bifurcation occurs: the baseline without VAPR plateaus and subsequently degrades, whereas the VAPR-enabled policy continues its upward trajectory. \Cref{fig: VAPR_kl} further reveals that each policy reset effectively extinguishes accumulated regularization pressure, preventing the KL term from dominating the gradients. Consistently, \cref{tab:ablation_settings} confirms that VAPR yields higher scores across all metrics (row 1 \vs row 4), validating its role in sustaining long-term optimization.


\begin{table}[!tb]
  \centering
  \scriptsize
\captionsetup{justification=raggedright,singlelinecheck=false}
  \caption{Ablation study of three-stage training pipeline on MIMIC\mbox{-}CXR. NLG Metrics: BLEU-1 to BLEU-4 (B-1--4), ROUGE-L (R-L), and METEOR (MTR). CE Metrics: CheXpert precision (P), CheXpert recall (R), CheXpert F1 (F1), and RadGraph F1 (RG). $\mathrm{\Delta}_{\text{NLG}}$ and $\mathrm{\Delta}_{\text{CE}}$ denote gains over Stage~1 (S1) for NLG and CE metrics.}
  \label{tab:ablation_stages}
  \setlength{\tabcolsep}{2.5pt}
  \renewcommand{\arraystretch}{1.10}

  \begin{tabular}{*{3}{>{\centering\arraybackslash}p{0.4 cm}}|
    *{6}{S[table-format=1.3]}|
    c|
    *{4}{S[table-format=1.3]}|
    c
  }
  \toprule
    \boldres{S1} & 
    \boldres{S2} &
    \boldres{S3} &
    \boldres{B-1} &
    \boldres{B-2} &
    \boldres{B-3} &
    \boldres{B-4} &
    \boldres{R-L} & \boldres{MTR} & \boldres{$\mathrm{\Delta}_{\text{NLG}}$} &
    \boldres{P} & \boldres{R} & \boldres{F1} &
     \boldres{RG} & \boldres{$\mathrm{\Delta}_{\text{CE}}$} \\
    \midrule
    \checkmark & & &
    0.383 & 0.236 & 0.157 & 0.109 &
    0.268 & 0.149 & \text{--} & 0.480
     & 0.353 & 0.407 & 0.215 & \text{--} \\
    \checkmark & \checkmark & &
    0.401 & 0.251 & 0.167 & 0.118 &
    0.277 & 0.156 & +5.2\% &
    0.526 & 0.431 & 0.474 & 0.220& +13.5\% \\
    
    \checkmark & & \checkmark &
    0.407 & 0.261 & 0.179 & 0.128 &
    0.288 & 0.157 & +9.1\% &
    0.516 & 0.393 & 0.446 & 0.228 & +8.8\% \\

    & \checkmark & \checkmark &
    0.415 & 0.271 & 0.185 & 0.131 &
    0.289 & 0.164 & +11.8\% &
    0.496 & 0.467 & 0.481 & 0.242 & +15.9\% \\
    
    \checkmark & \checkmark & \checkmark &
    0.424 & 0.280 & 0.195 & 0.143 &
    0.293 & 0.170 & +15.6\% &
    0.569 & 0.455 & 0.506 & 0.246 & +22.1\% \\
    \bottomrule
  \end{tabular}
\end{table}

\begin{table}[!tb]
  \centering
  \scriptsize
  \captionsetup{justification=raggedright,singlelinecheck=false}
  \caption{Ablation study of RL\mbox{-}stage modifications on MIMIC-CXR. NLG Metrics: BLEU-1 to BLEU-4 (B-1--4), ROUGE-L (R-L), and METEOR (MTR). CE Metrics: CheXpert precision (P), CheXpert recall (R), CheXpert F1 (F1), and RadGraph F1 (RG). Abbreviations: RER = Response-Weighted Regularization; CKL = constant KL; VAPR = Validation-Anchored Policy Reset. Best scores are in bold.}
  \label{tab:ablation_settings}
  \setlength{\tabcolsep}{4.2pt}
  \renewcommand{\arraystretch}{1.10}

  \sisetup{table-number-alignment = center,detect-weight=true,detect-family=true}
  \begin{tabular}{
    >{\raggedright\arraybackslash}p{2.5cm}|
    *{6}{S[table-format=1.3]} | *{4}{S[table-format=1.3]}
  }
  \toprule
    \boldres{Modifications} & \boldres{B-1} & \boldres{B-2} & \boldres{B-3} & \boldres{B-4} & \boldres{R-L} & \boldres{MTR} &
    \boldres{P} & \boldres{R} & \boldres{F1} & \boldres{RG}\\
    \midrule
    Ours (RER+VAPR)   & \boldres{0.424} & \boldres{0.280} & \boldres{0.195} & \boldres{0.143} & 0.293 & \boldres{0.170} &
    \boldres{0.569} & \boldres{0.455} & \boldres{0.506} & \boldres{0.246}\\
    w/ CKL, w/o RER         & 0.419 & 0.276 & 0.192 & 0.138 & \boldres{0.294} & 0.168 &  0.528 & 0.448 & 0.484 & 0.239 \\
    w/o KL                  & 0.414 & 0.267 & 0.184 & 0.132 & 0.290 & 0.165 & 0.522 & 0.416 & 0.463 & 0.235\\
    w/o VAPR                 & 0.417 & 0.272 & 0.187 & 0.134 & 0.289 & 0.167 & 0.530 & 0.439 & 0.481 & 0.241\\
    \bottomrule
  \end{tabular}
\end{table}

\captionsetup[subfigure]{justification=centering,singlelinecheck=false,font=small,skip=2pt}

\subsection{Qualitative Results}
\begin{figure*}[!t]
  \centering
   \includegraphics[width=0.99\linewidth]{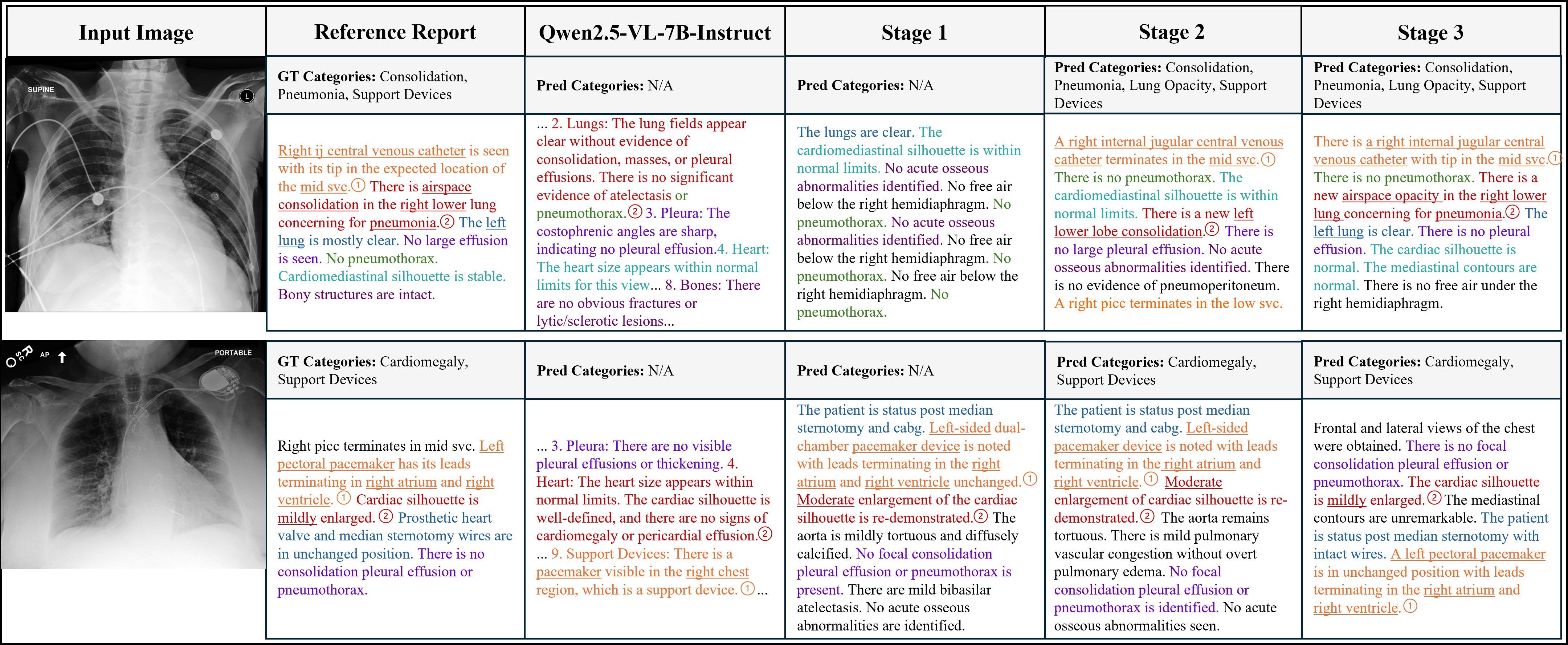}
\caption{Qualitative comparison examples on MIMIC-CXR: baseline (Qwen2.5-VL-7B-Instruct) and Stage~1 to Stage~3. Columns: input image; reference with ground-truth (GT) categories; and outputs with classifier-predicted categories. Sentences referring to the same anatomical region are color-coded and key phrases are underlined.}

\label{fig:qualitative}

   \label{fig: qualitative}
\end{figure*}

\Cref{fig: qualitative} compares outputs across the three stages and the baseline (Qwen2.5-VL-7B-Instruct without fine-tuning). Adding the classifier in Stage~2 helps identify clinically relevant cues that the baseline and Stage~1 miss. RL in Stage~3 improves wording precision and reduces hallucinated content.

\boldres{Case 1.} The baseline and Stage~1 miss lung abnormalities. With disease classifier guidance, Stage~2 identifies \textit{lobar consolidation} but assigns it to the wrong location (\textit{left lower}) and uses vague phrasing. Stage~3 corrects these issues: it describes the finding as \textit{airspace opacity}, localizes it to the \textit{right lower lung}, and infers \textit{pneumonia}, consistent with the reference.

\boldres{Case 2.} The baseline detects the \textit{pacemaker} but misses heart enlargement. Stage~1 and Stage~2 overestimate cardiac size as \textit{moderate} enlargement and introduce several hallucinations, including \textit{tortuous aorta}, \textit{atelectasis}, and \textit{mild pulmonary vascular congestion}. Stage~3 removes all hallucinations and correctly describes the cardiac silhouette as \textit{mildly} enlarged.

\section{Conclusion}
We present REVA-PO, a policy optimization framework for LVLM-based CXR report generation. By integrating Response-Weighted Regularization (RER) with Validation-Anchored Policy Reset (VAPR), REVA-PO enhances training stability and exploratory capacity. RER dynamically scales response-level penalties via advantages and reference entropy, while VAPR periodically resets policies to the best-performing anchor to alleviate accumulated regularization pressure. Evaluations on MIMIC-CXR and IU-Xray confirm that REVA-PO achieves state-of-the-art performance in both linguistic fluency and clinical accuracy.

\section*{Acknowledgements}
We thank Dr. Jianfeng Wang from the Department of Radiology at The First People's Hospital of Zhengzhou for his assistance with the radiological evaluation in this study. We also acknowledge UBC Advanced Research Computing and the Digital Research Alliance of Canada for providing the computational resources that supported this work.

%
%
\bibliographystyle{splncs04}
\bibliography{main}
\clearpage
\appendix
\section*{Supplementary Material}
\input{supplementary}
\end{document}

%% file: supplementary.tex
\renewcommand{\thetable}{S\arabic{table}}
\renewcommand{\thefigure}{S\arabic{figure}}
\section{Reward Selection and Analysis}
To optimize the reinforcement learning (RL) process, we evaluate three candidate metrics as proxy rewards: \boldres{BLEU-4} \cite{papineni2002bleu}, \boldres{CheXpert F1} \cite{irvin2019chexpert}, and \boldres{RadGraph F1} \cite{delbrouck2024radgraph}. BLEU-4 assesses local linguistic coherence via 4-gram overlaps, while CheXpert F1 and RadGraph F1 focus on clinical accuracy—measuring diagnostic labels and the structural alignment of clinical entities and their relations, respectively. Although CheXpert F1 and RadGraph F1 seem intuitively superior for aligning models with clinical ground truth, they exhibit significant limitations when used as direct optimization objectives in RL. For the reasons discussed below, we therefore adopt BLEU-4 as our proxy reward.

\subsection{CheXpert F1}
\boldres{High Reward Variance.} Using the policies both before and after RL, we sample $G=4$ candidate responses $\{o_{i,j}\}_{j=1}^{G}$ for each prompt $q_i$ in the MIMIC-CXR validation set $\mathcal{D}_{\mathrm{val}}$ (where $q_i \sim \mathcal{D}_{\mathrm{val}}$ and $|\mathcal{D}_{\mathrm{val}}|=2130$). We then calculate BLEU-4, CheXpert F1, and RadGraph F1 scores for each response and compute the standard deviation within each group. As illustrated in \cref{fig:std}, CheXpert F1 exhibits high variance both before and after RL. This volatility produces high-variance gradients, leading to training instability. In our experiments, using CheXpert F1 as the reward consistently triggered policy collapse during the early stages of training.

\subsection{RadGraph F1}

\begin{figure*}[!tb]
  \centering
  \begin{subfigure}[t]{0.49\textwidth}
    \centering
    \includegraphics[width=\linewidth]{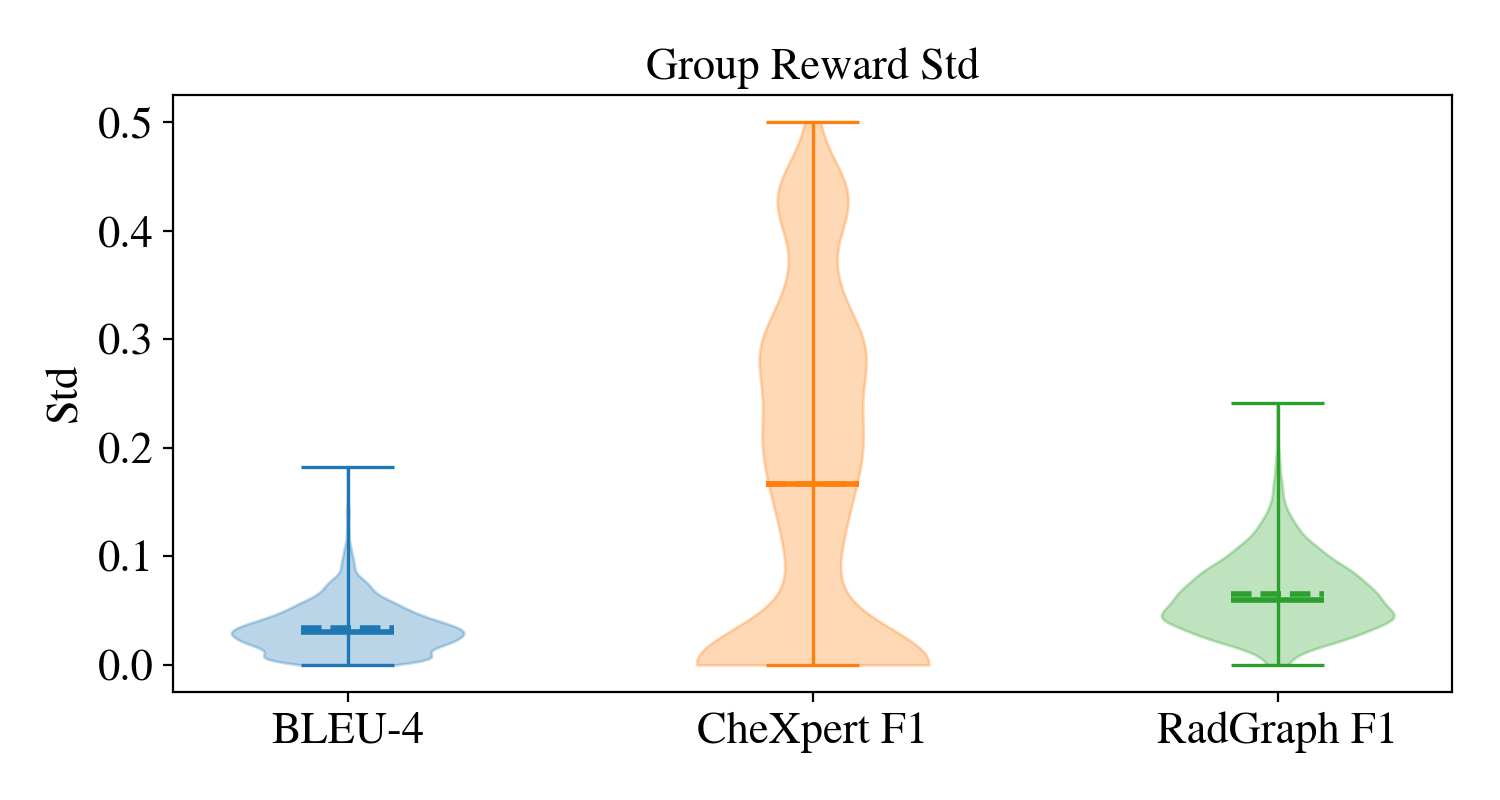}
    \subcaption{Before RL}
    \label{fig:s2_std}
  \end{subfigure}
  \begin{subfigure}[t]{0.49\textwidth}
    \centering
    \includegraphics[width=\linewidth]{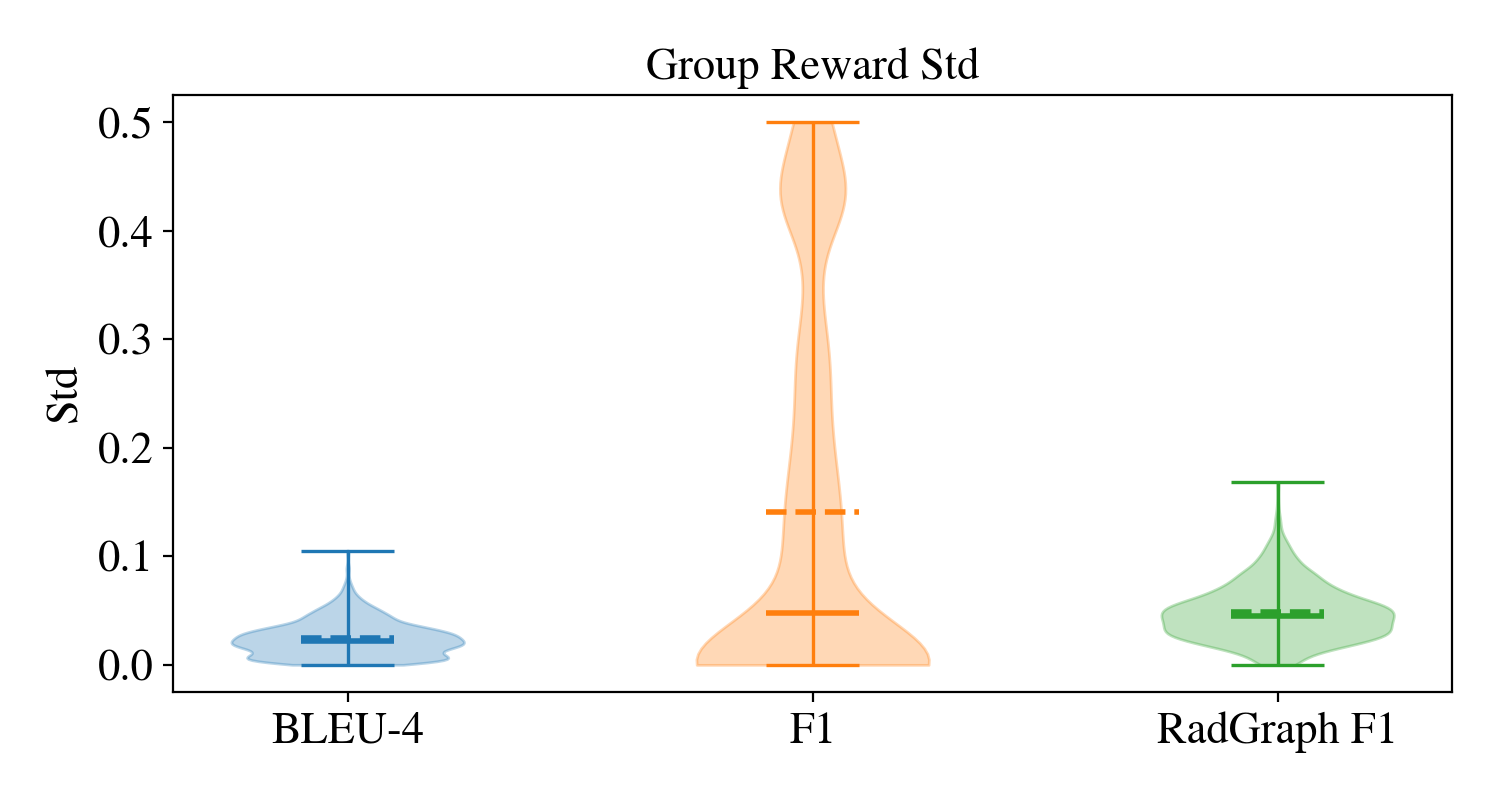}
    \subcaption{After RL}
    \label{fig:std_s3}
  \end{subfigure}
  \caption{
  Distribution of within-group standard deviations (std) for different rewards. Solid lines represent the maximum, median, and minimum values, respectively, while dashed lines represent the mean.
  }
  \label{fig:std}
\end{figure*}

\begin{figure*}[!tb]
  \centering
    \includegraphics[width=\linewidth]{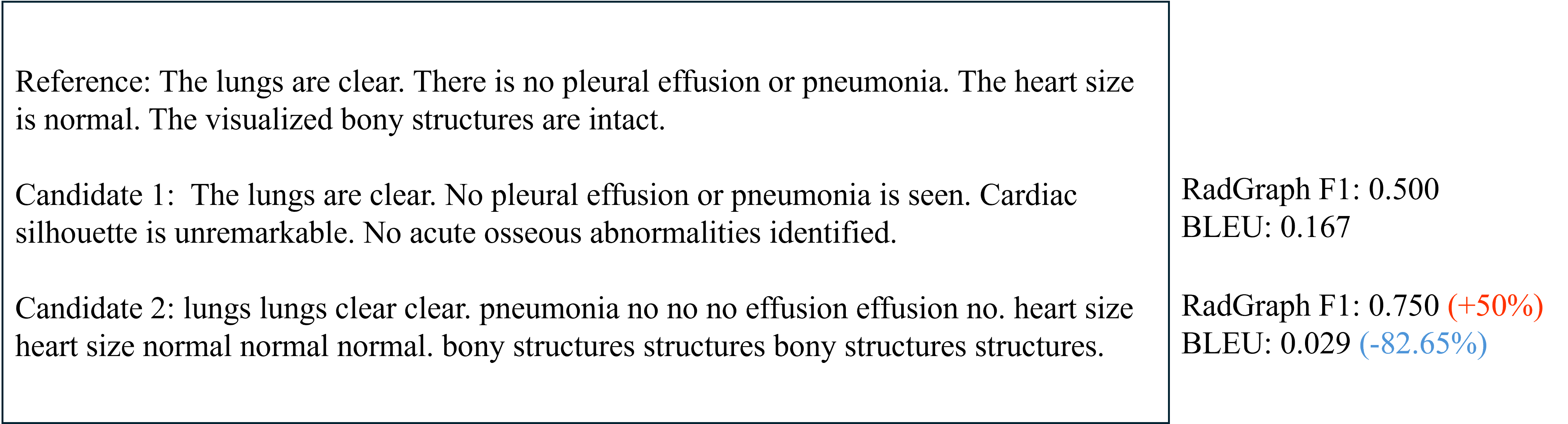}
  \caption{
  RadGraph F1 can exhibit a reward inversion phenomenon, where the lower-quality report receives a higher reward score.
  }
  \label{fig:reward_inversion}
\end{figure*}

\boldres{Reward Inversion.} Although \cref{fig:std} shows that RadGraph F1 is relatively stable, it suffers from a different critical issue: reward inversion. Because RadGraph F1 relies on sparse, entity-centric signals, it makes the reward function susceptible to hacking via grammatically incoherent entity stuffing. RadGraph's entity extractor prioritizes isolated clinical nodes (entities and relations) over syntactic integrity, allowing nonsensical reports to achieve inflated scores simply by concatenating high-frequency medical keywords. As demonstrated in \cref{fig:reward_inversion}, Candidate 2 achieves a higher RadGraph F1 reward despite offering significantly lower clinical utility and readability than Candidate 1. This inversion encourages the policy to sacrifice narrative logic to maximize the proxy score. Conversely, BLEU-4 provides denser, token-level supervision and implicitly favors structural coherence through continuous $n$-gram matching. Consequently, as shown in \cref{tab:ablation_rewards}, while using RadGraph F1 as the reward improves its own metric, performance on other NLG and CE metrics remains distinctly inferior to models trained with BLEU-4.

\subsection{BLEU-4 Improves Clinical Efficacy}
We further investigate the empirical alignment between BLEU-4 and clinical metrics to justify its use as a surrogate reward. Specifically, we analyze the consistency of the advantage signals generated by BLEU-4 and CheXpert F1. For a set of the sampled responses $\{\{o_{i,j}\}_{j=1}^{G}\}_{i=1}^{|\mathcal{D}_{\mathrm{val}}|}$, we compute the group-relative advantage $A_{i,j}$ for each metric. Since the sign of $A_{i,j}$ dictates the direction of policy updates (i.e., whether to increase or decrease the log-probability of a response), sign consistency implies alignment in the optimization gradient. We define the Sign Consistency Probability $P_{\mathrm{sign}}$ as 
\begin{equation}
    P_{\mathrm{sign}}^\pi = \frac{1}{|\mathcal{D}_{\mathrm{val}}| \cdot G} \sum_{i=1}^{|\mathcal{D}_{\mathrm{val}}|} \sum_{j=1}^{G} \mathbb{I} \left( \text{sign}(A_{i,j}^{\mathrm{BLEU}}) = \text{sign}(A_{i,j}^{\mathrm{F1}}) \right),
\end{equation}
and the Top-1 Selection Consistency $P_{\mathrm{top1}}$ as:
\begin{equation}
P_{\mathrm{top1}}^\pi = \frac{1}{|\mathcal{D}_{\mathrm{val}}|} \sum_{i=1}^{|\mathcal{D}_{\mathrm{val}}|} \mathbb{I} \left( \arg\max_{j} R_{i,j}^{\mathrm{BLEU}} = \arg\max_{j} R_{i,j}^{\mathrm{F1}} \right),
\end{equation}
where $\mathbb{I}$ is the indicator function.

Our results show high initial alignment before RL ($\pi_{\mathrm{base}}$), with $P_{\mathrm{sign}}^{\pi_{\mathrm{base}}} = 0.756$ and $P_{\mathrm{top1}}^{\pi_{\mathrm{base}}} = 0.619$. This suggests that even as a purely linguistic metric, BLEU-4 inherently captures optimization directions that favor clinical accuracy. After RL optimization ($\pi_{\mathrm{RL}}$), these probabilities increase to $P_{\mathrm{sign}}^{\pi_{\mathrm{RL}}} = 0.805$ (+6.5\%) and $P_{\mathrm{top1}}^{\pi_{\mathrm{RL}}} = 0.782$ (+26.3\%). The significant rise in $P_{\mathrm{top1}}$ indicates that the policy has converged toward a region where high-quality linguistic structure and diagnostic correctness are strongly coupled. This alignment is further corroborated by the results in Tab. 3 of the main paper, where optimizing for BLEU-4 (Stage 3) consistently yields improvements in CE metrics, empirically validating BLEU-4 as an effective surrogate for clinical diagnostic accuracy.

\begin{table}[!tb]
  \centering
  \scriptsize
  \captionsetup{justification=raggedright,singlelinecheck=false}
  \caption{Ablation study of different rewards on MIMIC-CXR. NLG Metrics: BLEU-1 to BLEU-4 (B-1--4), ROUGE-L (R-L), and METEOR (MTR). CE Metrics: CheXpert precision (P), CheXpert recall (R), CheXpert F1 (F1), and RadGraph F1 (RG). Best scores are in bold.}
  \label{tab:ablation_rewards}
  \setlength{\tabcolsep}{4pt}
  \renewcommand{\arraystretch}{1.10}

  \sisetup{table-number-alignment = center,detect-weight=true,detect-family=true}
  \begin{tabular}{
    >{\raggedright\arraybackslash}p{1.7cm}|
    *{6}{S[table-format=1.3]} | *{4}{S[table-format=1.3]}
  }
    \toprule
    \boldres{Rewards} & \boldres{B-1} & \boldres{B-2} & \boldres{B-3} & \boldres{B-4} & \boldres{R-L} & \boldres{MTR} &
    \boldres{P} & \boldres{R} & \boldres{F1} & \boldres{RG}\\
    \midrule
    Before RL                & 0.401 & 0.251 & 0.167 & 0.118 & 0.277 & 0.156 & 0.526 & 0.431 & 0.474 & 0.220\\
    RadGraph F1                & 0.414 & 0.262 & 0.175 & 0.123 & 0.281 & 0.165 & 0.529 & 0.440 & 0.480 & \boldres{0.259}\\
    BLEU-4  & \boldres{0.424} & \boldres{0.280} & \boldres{0.195} & \boldres{0.143} & \boldres{0.293} & \boldres{0.170} &
    \boldres{0.569} & \boldres{0.455} & \boldres{0.506} & 0.246\\
    \bottomrule
  \end{tabular}
\end{table}

\section{Computational Overhead of Response-Weighted Regularization}
While our method does not require additional architectural components or preference data, it does introduce a slight computational overhead. This occurs because calculating the reference-policy entropy (Eq. 7) necessitates averaging over the entire vocabulary. Empirically, on the MIMIC-CXR dataset, transitioning to RER (constant $\beta$ $\rightarrow$ response-weighted $\beta_i$) increases the average step time from 11.80 s to 12.02 s (+1.9\%) and peak memory usage from 21.62 GiB to 22.46 GiB (+3.9\%). Ultimately, this marginal increase in computational cost is small and remains negligible compared to the substantial overhead associated with deploying a separate critic-based classifier.

\section{Prompt Format}
\label{sec:iprompt_format}
As shown in \cref{fig:prompt-example}, Stage~1 and Stages~2 \& 3 use
prompts $q = [I, V]$ and $q = [I, D, V]$, respectively, where $I$ denotes
instruction tokens, $V$ denotes image tokens, and $D$ denotes disease
label tokens predicted by the classifier. For each sample, $V$ and $D$ vary across images, whereas $I$ is fixed.

\begin{figure}[tb]
  \centering
  \small

  \fbox{%
    \parbox{0.95\linewidth}{%
      \boldres{Stage 1}\\[4pt]
      \boldres{Prompt:}\quad
      \texttt{[image token]} \dots \texttt{[image token]}
      You are a radiologist. Analyze the input chest X\mbox{-}ray. Assess the major
      anatomical regions if visible (e.g., airways, lungs, pleura, heart,
      mediastinum, great vessels, diaphragm, bones, upper abdomen
      such as liver and stomach, and support devices). Describe any
      observed findings that are clinically relevant, whether normal
      or abnormal.
    }%
  }

  \vspace{6pt}

  \fbox{%
    \parbox{0.95\linewidth}{%
      \boldres{Stage 2 and Stage 3}\\[4pt]
      \boldres{Prompt:}\quad
      \texttt{[image token]} \dots \texttt{[image token]}
      You are a radiologist. Analyze the input chest X\mbox{-}ray. Assess the major
      anatomical regions if visible (e.g., airways, lungs, pleura, heart,
      mediastinum, great vessels, diaphragm, bones, upper abdomen
      such as liver and stomach, and support devices). Describe any
      observed findings that are clinically relevant, whether normal
      or abnormal. \\
      The following candidate findings may be present:
      [\texttt{Cardiomegaly}, \texttt{Lung Opacity}].
    }%
  }

  \caption{Prompt examples for Stage 1, Stage 2, and Stage 3 on MIMIC-CXR.}
  \label{fig:prompt-example}
\end{figure}

\section{ Implementation Details}
\subsection{Preliminary Study}
In a preliminary study, we compared two widely used open-source LVLMs,
Qwen2.5-VL-7B-Instruct~\cite{bai2025qwen2} and
Llama-3.2-11B-Vision-Instruct~\cite{grattafiori2024llama}.
All preliminary experiments were run on 8 NVIDIA A100 40GB GPUs, which
support both bfloat16 (BF16) and float16 (FP16).
In Stage~1 and Stage~2, the two models achieved similar performance.
Under identical settings, Llama required more memory and was stable only with
BF16; training with FP16 caused gradient explosions.
Qwen required less memory and trained stably with both BF16 and FP16.
We therefore adopt Qwen2.5-VL-7B-Instruct as our main baseline.
To study the effect of parameter count, we also include
Qwen2.5-VL-3B-Instruct as a 3B baseline.
All reported experiments are run on 8 NVIDIA Tesla V100 (32GB) GPUs with FP16.

\subsection{Hyperparameters}
We denote batch size by $bs$, learning rate by $lr$, weight decay by $wd$, and the number of training epochs by $E$.
We use DeepSpeed ZeRO\footnote{https://github.com/deepspeedai/DeepSpeed} stage~1 to reduce the GPU memory required for training.

\boldres{Stage 1 Warm-up.}
Only the merger module is trainable, and all other components are frozen. Following established practices for training only the merger module \cite{liu2023visual, liu2024improved}, we set the learning rate to $1 \times 10^{-3}$. For MIMIC-CXR, we set $bs = 32$, and $E = 10$.
We use the AdamW optimizer with $\beta_1 = 0.9$ and $\beta_2 = 0.999$, and set the $wd=1\times10^{-5}$.
For IU-Xray, we initialize the merger using weights pre-trained during Stage 1 on MIMIC-CXR. We set $bs = 32$, $E = 3$, and $wd = 1\times10^{-4}$.

\boldres{Fine-tuning Classifier.} We use CheXpert\footnote{https://github.com/stanfordmlgroup/chexpert-labeler} to extract disease categories from each report.
Specifically, for each report, CheXpert outputs the presence of 14 categories
(no finding, enlarged cardiomediastinum, cardiomegaly, lung opacity,
lung lesion, edema, consolidation, pneumonia, atelectasis, pneumothorax,
pleural effusion, pleural other, fracture, and support devices).
Each category is predicted as $1$, $0$, $-1$, or empty, representing
positive, negative, uncertain, and not mentioned, respectively.
We take the positive categories as the label set for the corresponding chest
X\mbox{-}ray.
Due to the inherent noise in CheXpert, a few reports have no positive
categories (and no finding is not marked as positive).
For these reports, we manually add no finding so that each image has
at least one label and the classifier can be trained correctly.

The disease classifier is initialized from Medical MAE~\cite{xiao2023delving} ViT-base/16, takes
chest X\mbox{-}rays as input, and outputs the 14 labels.
For MIMIC\mbox{-}CXR, we set $bs = 1600$, $lr = 1.5\times10^{-3}$,
$E = 20$, $wd= 0.05$, and $\beta_2 = 0.95$; without reducing
$\beta_2$, this large batch size caused training to crash.
For IU\mbox{-}Xray, we set $bs = 64$, $lr = 6.25\times10^{-5}$,
$E = 20$, $wd = 0.05$, and $\beta_2 = 0.999$.
Because IU\mbox{-}Xray is small, we initialize the classifier from the model
fine-tuned on MIMIC\mbox{-}CXR.
The classifier remains frozen in Stages~2 and~3.

\boldres{Stage 2 Classifier-Guided SFT.} For the LVLM, we train the merger and apply LoRA \cite{hu2022lora} to all other linear layers. The LoRA rank $r$ and $\alpha$ are both 32, and the LoRA dropout is 0.1. We set $bs=28$, $lr=1\times10^{-4}$, and $wd= 1\times10^{-3}$. For MIMIC-CXR, we use $E=1$; for IU-Xray, we use $E=5$.

\boldres{Stage 3 RL.} The set of trainable parameters and the LoRA hyperparameters are the same as in Stage~2. Because RL training is very slow (for example, one epoch on the full MIMIC\mbox{-}CXR training set with 271k samples takes approximately 5 days on 8 V100 GPUs), we train on a subset of MIMIC\mbox{-}CXR for Stage 3.
Specifically, after tokenizing the reports in the training set, we discard reports that are shorter than 30 tokens, which contain overly simplistic descriptions that provide insufficient supervision signals. From the remaining reports, we randomly sample 10k examples as the RL subset. The validation and test sets remain unchanged. The IU\mbox{-}Xray training set contains only 2k samples, so we do not subsample it.

We set $bs = 8$, $lr = 3\times10^{-6}$, and $wd = 1\times10^{-3}$.
For MIMIC\mbox{-}CXR, we use $E = 9$; for IU\mbox{-}Xray, we use $E = 15$.
Since the 3B baseline converges more slowly, we use $E = 12$ on MIMIC\mbox{-}CXR for 3B model.

For each prompt, the old policy $\pi_{\theta_{\mathrm{old}}}$ generates $G=4$ candidate reports. For $\mathcal{J}_{\mathrm{REVA-PO}}$ objective, the clipping parameters are set to $(\epsilon_{\text{low}}, \epsilon_{\text{high}}) = (0.2, 0.28)$. For MIMIC-CXR, the KL bounds are $(\beta_{\min}, \beta_{\mathrm{base}}, \beta_{\max}) = (0.05, 0.10, 0.12)$ and balancing coefficient is $\lambda = 0.1$. For IU\mbox{-}Xray, KL bounds are $(0.05, 0.10, 0.15)$, and $\lambda$ is $0.1$; an entropy coefficient of $0.001$ is also applied. The old policy update period is $N = 16$ steps, and the reference update period is $M = 3$ epochs.

\section{Impact of Group Size on Stability}
\begin{figure}[t]
\centering

\begin{tabular}{@{}p{0.66\columnwidth}@{\hspace{0.02\columnwidth}}p{0.32\columnwidth}@{}}

\begin{minipage}[t]{\linewidth}
  \vspace{0pt}\centering
  \begin{subfigure}[t]{0.49\linewidth}
    \centering
    \includegraphics[width=\linewidth]{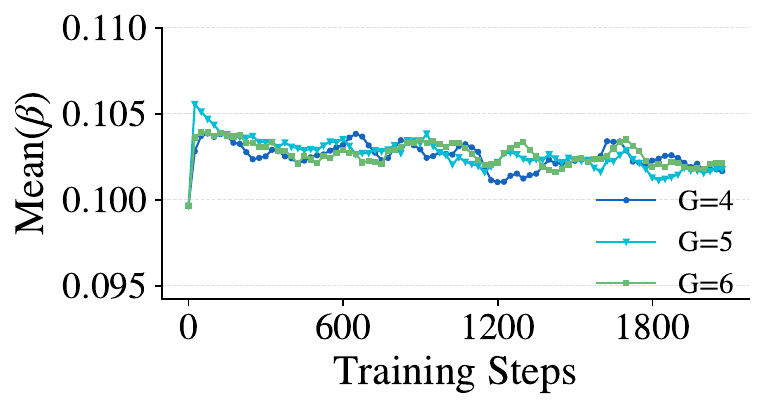}
  \end{subfigure}\hfill
  \begin{subfigure}[t]{0.49\linewidth}
    \centering
    \includegraphics[width=\linewidth]{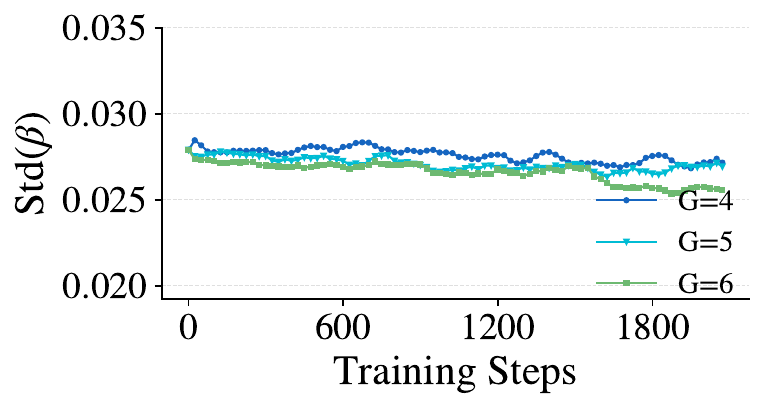}
  \end{subfigure}
\end{minipage}
&
\begin{minipage}[t]{\linewidth}
  \vspace{0pt}\centering
\scriptsize
\captionsetup{justification=raggedright,singlelinecheck=false} 
\label{tab:7x5_singlecol}
\setlength{\tabcolsep}{1.7pt} \renewcommand{\arraystretch}{1.10} 
\sisetup{mode=text,table-number-alignment=center,detect-weight=true,detect-family=true} 
\begin{tabular}{ >{\centering\arraybackslash}p{0.3cm}| S[table-format=1.3] S[table-format=1.3] S[table-format=1.3] S[table-format=1.3] } \toprule \boldres{$G$} & \boldres{B-1} & \boldres{B-4} & \boldres{R-L} & \boldres{MTR} \\ \midrule 4 & 0.514 & \boldres{0.199} & 0.403 & 0.213 \\ 5 & 0.511 & 0.196 & 0.401 & 0.210 \\ 6 & \boldres{0.517} & \boldres{0.199} & \boldres{0.405} & \boldres{0.215} \\ \bottomrule \end{tabular} 
\end{minipage}
\\[-2pt] 

\begin{minipage}[t]{\linewidth}
  \vspace{0pt}
  \captionsetup{skip=0pt}
  \captionof{figure}{Mean and std of $\beta$ on IU\mbox{-}Xray for $G\in\{4,5,6\}$.}
  \label{fig:beta}
\end{minipage}
&
\begin{minipage}[t]{\linewidth}
  \vspace{0pt}
  \captionsetup{skip=0pt}
  \captionof{table}{Stage~3 results on IU\mbox{-}Xray for $G\in\{4,5,6\}$.}
  \label{fig:results}
\end{minipage}

\end{tabular}
\end{figure}

In this section, we evaluate the impact of increasing the group size $G$ on training stability and model performance. Because a NVIDIA V100 32GB GPU is memory-constrained to a maximum of $G=4$, the experiments in this ablation study are conducted using 8 NVIDIA A100 (40GB) GPUs. As illustrated in \cref{fig:beta}, for $G \in \{4, 5, 6\}$, the response-weighted $\beta_i$ generated by the RER mechanism remains consistently stable throughout training. Furthermore, increasing $G$ has minimal impact on the final evaluation metrics, although setting $G=6$ yields a slight performance improvement as shown in \cref{fig:results}. These findings demonstrate that our proposed RER is highly robust to different group sizes.

\section{Impact of Model Size (3B vs 7B)}
As shown in \cref{tab:supp_results}, REVA-PO-7B consistently outperforms REVA-PO-3B on both datasets. On the smaller IU-Xray dataset, the performance gap between the 3B and 7B models is small, whereas on the larger and more complex MIMIC-CXR dataset, the 7B model has a clear advantage. On IU-Xray, the 3B model matches the 7B model on BLEU-2 and METEOR, is slightly lower on BLEU-1, BLEU-3, and BLEU-4, and shows a clear decrease on ROUGE-L. On MIMIC-CXR, REVA-PO-7B exceeds REVA-PO-3B across all metrics.

These results suggest that on a smaller dataset, models with different parameter counts can reach similar performance, whereas on a larger dataset, a larger model yields clear gains. We also observe that on MIMIC-CXR, REVA-PO-3B converges more slowly than REVA-PO-7B during the RL stage.

\begin{table}[!t]
  \centering
 \scriptsize
\captionsetup{justification=raggedright,singlelinecheck=false}
  \caption{Comparisons between REVA-PO-3B and REVA-PO-7B on IU-Xray and MIMIC-CXR using BLEU-1 to BLEU-4 (B-1--4), ROUGE-L (R-L), and METEOR (MTR). REVA-PO-3B and REVA-PO-7B use Qwen2.5-VL-3B-Instruct and Qwen2.5-VL-7B-Instruct as baselines, respectively. Best scores are in bold.}
  \label{tab:supp_results}
  \setlength{\tabcolsep}{5pt}
  \renewcommand{\arraystretch}{1.10}

  \begin{tabular}{>{\centering\arraybackslash}m{1.65cm}|>{\centering\arraybackslash}p{2cm}|*6{S[table-format=1.3]}}
    \toprule
    \boldres{Datasets} & \boldres{Methods} &
    \boldres{B-1} & \boldres{B-2} & \boldres{B-3} & \boldres{B-4} & \boldres{R-L} & \boldres{MTR} \\
    \midrule

    \multirow{2}{*}{\boldres{IU-Xray}}
      
      & REVA-PO-3B                 & 0.510 & \boldres{0.356} & 0.260 & 0.195 & 0.392 & \boldres{0.216} \\
      & REVA-PO-7B                 & \boldres{0.517} & \boldres{0.356} & \boldres{0.261} & \boldres{0.199} & \boldres{0.406} & \boldres{0.216} \\
    \midrule

    \multirow{2}{*}{\boldres{MIMIC-CXR}}
      & REVA-PO-3B                 & 0.412 & 0.272 & 0.187 & 0.135 & 0.290 & 0.163 \\
      & REVA-PO-7B                 & \boldres{0.424} & \boldres{0.280} & \boldres{0.195} & \boldres{0.143} & \boldres{0.293} & \boldres{0.170} \\
    \bottomrule
  \end{tabular}
\end{table}

\section{Classifier Performance}
\Cref{tab:cls_performance} details the performance of the employed disease classifier on MIMIC-CXR. A key advantage of our REVA-PO framework is its strictly classifier-agnostic design. Consequently, as more advanced clinical classifiers are developed in the future, they can be seamlessly integrated as plug-and-play replacements. Updating the system to leverage these improved models would require retraining only Stages 2 and 3.
\begin{table}[t]
\centering
\scriptsize
\setlength{\tabcolsep}{6pt}
\renewcommand{\arraystretch}{1.10}
\caption{Disease classifier performance on MIMIC\mbox{-}CXR.}
\label{tab:cls_performance}
\begin{tabular}{lc|lc|lc}
\toprule
\boldres{Category} & \boldres{AUC} &
\boldres{Category} & \boldres{AUC} &
\boldres{Category} & \boldres{AUC} \\
\midrule
No Finding        & 0.809 &
Atelectasis       & 0.775 &
Lung Lesion       & 0.762 \\
Enl.\ Cardiom.    & 0.655 &
Pneumothorax      & 0.829 &
Edema             & 0.839 \\
Pleural Effusion  & 0.902 &
Consolidation     & 0.788 &
Lung Opacity      & 0.736 \\
Pleural Other     & 0.816 &
Pneumonia         & 0.706 &
Fracture          & 0.749 \\
Support Devices   & 0.913 &
Cardiomegaly      & 0.789 &
Average & 0.873 \\
\bottomrule
\end{tabular}%
\end{table}

\section{Details of SOTA Methods}
\Cref{tab:additional_resource} compares the LLM sizes and additional training resources of state-of-the-art (SOTA) methods beyond the standard IU-Xray and MIMIC-CXR datasets. While most SOTA methods rely on costly auxiliary resources or alter the evaluation data, our approach achieves SOTA performance using only standard image-level labels while strictly preserving the integrity of the validation and test sets.

Specifically, MambaXray-VL~\cite{wang2025cxpmrg} and B-LLM~\cite{liu2024bootstrapping} require massive external datasets, whereas STREAM~\cite{yang2025spatio}, CoD~\cite{jin2025chain}, and RGRG~\cite{tanida2023interactive} depend on expensive dense annotations (\eg, bounding boxes). DCL~\cite{li2023dynamic} relies on external knowledge graphs. Furthermore, several methods alter the evaluation protocol: HERGen~\cite{wang2024hergen} removes lateral images, while CMCRL~\cite{chen2025cross} and MA~\cite{shen2024automatic} truncate reports. By avoiding such data filtering and truncation, our method ensures fairer and more straightforward future comparisons. Only PromptMRG~\cite{jin2024promptmrg}, Clinical-BERT~\cite{yan2022clinical}, and AM-MRG~\cite{11142850} match our resource-efficient setting of utilizing solely image-level classification labels.

\begin{table*}[!t]
  \centering
\scriptsize
\captionsetup{justification=raggedright,singlelinecheck=false}
    \caption{Large language models and additional training resources used by state-of-the-art methods. Here, additional training resources denote any supervision beyond the original image–report pairs in IU-Xray and MIMIC-CXR, including extra datasets, fine-grained annotations, knowledge graphs, or modified reports and data splits. Most prior methods use LLMs of similar or smaller size than REVA-PO but rely on such extra supervision, whereas REVA-PO uses only image-level classification labels and preserves all evaluation samples.}
  \label{tab:additional_resource}
  \setlength{\tabcolsep}{3pt}
  \renewcommand{\arraystretch}{1.05}
  \begin{tabular}
  {>{\raggedright\arraybackslash}m{2.5cm}|
   >{\centering\arraybackslash}m{0.7cm}|
   >{\centering\arraybackslash}m{2cm}|
   >{\raggedright\arraybackslash}p{6.1cm}}
    \toprule
    \boldres{Methods} & \boldres{Years} &
    \boldres{Large Language Models} & \boldres{Additional Training Resources} \\
    \midrule
    Clinical-BERT \cite{yan2022clinical} &
    2022 & \text{--} & 1. Classification labels for each CXR. \\
    \midrule
    CMMRL \cite{qin2022reinforced} &
    2022 & \text{--} & \text{--} \\
    \midrule
    DCL \cite{li2023dynamic} & 2023 & \text{--} & 1. Chest knowledge graph~\cite{zhang2020radiology}\\
    \midrule
    NADM \cite{zhao2023normal} & 2023 & \text{--} & 1. Medical concepts extracted from each CXR by RadGraph~\cite{jain2021radgraph}\\
    \midrule
    METransformer \cite{wang2023metransformer} & 2024 & \text{--} & \text{--} \\
    \midrule
    PromptMRG \cite{jin2024promptmrg}       & 2024 & \text{--} & 1. Classification labels for each CXR. \\
    \midrule
    MA \cite{shen2024automatic}   & 2024 & \text{--} & \makecell[l]{1. Truncate each IU report to at most 60 words. \\ 2. Truncate each MIMIC report to at most 100 \\words.}\\
    \midrule
    CMCRL \cite{chen2025cross} & 2025 & \text{--} & \makecell[l]{1. Truncate each IU report to at most 60 words, \\and only tokenize words that occur more than 3 \\times.\\ 2. Truncate each MIMIC report to at most 80 \\words,
    and only tokenize words that occur more\\ than 10 times.}\\
    \midrule
    RGRG \cite{tanida2023interactive} & 2023 & GPT-2-Medium-355M~\cite{radford2019language} & \makecell[l]{1. Bounding boxes, region descriptions, region\\ labels, and abnormality classification labels for\\ 29 anatomical regions in each CXR of the Chest\\ ImaGenome dataset~\cite{wu2021chest} (242,072 samples).}\\
    \midrule
    HERGen \cite{wang2024hergen}            & 2024 & DistilGPT-2-82M~\cite{sanh2019distilbert}& \makecell[l]{1. Remove lateral images from the train/val/test\\ sets of MIMIC. \\
    2. Truncate each MIMIC report to at most 60 \\words.}\\
    \midrule
     B-LLM \cite{liu2024bootstrapping} & 2024 & Vicuna-13B~\cite{chiang2023vicuna}& 1. Public medical corpora including 3,000 documents from PubMed. \\
     \midrule
     CoD \cite{jin2025chain}                & 2025 & Llama-2-7B~\cite{touvron2023llama}&
     \makecell[l]{1. Question–answer (QA) pairs from \\MIMIC-Diff-VQA~\cite{hu2023expert}, where each QA pair \\specifies the presence, location, severity, and \\causal relation of each disease for each CXR.\\
     2. Bounding boxes and classification labels of\\ multiple diseases from RSNA~\cite{rsna-pneumonia-detection-challenge} (5,659 samples), \\VinDR-CXR~\cite{nguyen2022vindr} (3,706 samples), and private \\(15,555 samples) datasets.} \\
     \midrule
     STREAM \cite{yang2025spatio}   & 2025 & TinyLlama-1.1B~\cite{zhang2024tinyllama} & \makecell[l]{1. Bounding boxes, region descriptions, region \\labels, and abnormality classification labels for \\29 anatomical regions in each CXR of the Chest\\ ImaGenome dataset~\cite{wu2021chest} (242,072 samples).}\\
     \midrule
     MambaXray-VL \cite{wang2025cxpmrg}   & 2025 & Llama-2-7B~\cite{touvron2023llama}& \makecell[l]{1. 1.27 million CXR~\cite{wang2024pre}.\\
     2. CheXpert Plus dataset~\cite{chambon2024chexpert} (210k image-report \\pairs) for pre-training.}\\
     \midrule
     AM-MRG \cite{11142850} &2026&Llama-2-7B~\cite{touvron2023llama} & 1. Classification labels for each CXR.\\
     \midrule
     REVA-PO (ours) &&Qwen2.5-VL-7B-Instruct~\cite{bai2025qwen2} & 1. Classification labels for each CXR.\\
     \bottomrule
    \end{tabular}
    \end{table*}


%
%